\documentclass[10pt,twocolumn,letterpaper]{article}

\usepackage[pagenumbers]{cvpr} % To force page numbers, e.g. for an arXiv version

\definecolor{cvprblue}{rgb}{0.21,0.49,0.74}
\usepackage[pagebackref,breaklinks,colorlinks,allcolors=cvprblue]{hyperref}

\def\paperID{5} % *** Enter the Paper ID here
\def\confName{CVPR}
\def\confYear{2025}

\title{Cross-Modal Consistency Learning for Sign Language Recognition}

\author{Kepeng Wu$^1$, Zecheng Li$^1$, Hezhen Hu$^3$, Wengang Zhou$^{1,2\thanks{Corresponding author}}$, Houqiang Li$^{1,2}$\\
$^1$MoE Key Laboratory of Brain-inspired Intelligent Perception and Cognition, \\
University of Science and Technology of China
\\
$^2$Institute of Artificial Intelligence, Hefei Comprehensive National Science Center\\
$^3$University of Texas at Austin
\\
{\tt\small \{wukp, lizecheng23\}@mail.ustc.edu.cn} 
\\
{\tt\small \{zhwg, lihq\}@ustc.edu.cn, alexhu@utexas.edu}
}
\def\modelname{\mbox{CCL-SLR}\xspace}
\makeatletter
\renewcommand{\maketag@@@}[1]{\hbox{\m@th\normalsize\normalfont#1}}%
\makeatother

\makeatletter
\DeclareRobustCommand\onedot{\futurelet\@let@token\@onedot}
\def\@onedot{\ifx\@let@token.\else.\null\fi\xspace}

\def\modelname{\mbox{CCL-SLR}\xspace}

\makeatother

\def\BibTeX{{\rm B\kern-.05em{\sc i\kern-.025em b}\kern-.08em
    T\kern-.1667em\lower.7ex\hbox{E}\kern-.125emX}}

\newcommand{\Pmodal}{\ding{52} & }
\newcommand{\Rmodal}{ & \ding{52} }
\newcommand{\RPmodal}{\ding{52} & \ding{52}}

\usepackage{dsfont}
\usepackage{bm}
\usepackage{multirow}
\usepackage{multicol}
\usepackage{booktabs}
\usepackage{pifont}
\usepackage{bbding}
\usepackage{xspace}

\begin{document}
\maketitle
\begin{abstract}
% The pre-training paradigm plays a dominant role in the Isolated Sign Language Recognition (ISLR) task.
% The pre-training paradigm is a privileged direction in the Isolated Sign Language Recognition (ISLR) task.
% Existing pre-training strategies that rely on the compact yet effective Pose modality have achieved promising performance. 
% However, these approaches are limited by the accuracy and robustness of current pose estimators, consequently overlooking the rich visual cues present in the original RGB videos.
% To this end, we propose a novel framework called {Cross-modal consistency Learning Sign Language Recognition}~(\modelname), elaborating detailed expression and gesture information embedded in the RGB modality to fertilize sign language representation via a self-supervised paradigm.
% Concretely, we employ the contrastive learning to implement instance discrimination within and across modalities. 
% Furthermore, we design the Motion-aware Mask Augmentation and the Cross-modal Positive Mining to mitigate the distribution gap between modalities. 
% Extensive experiments on four ISLR benchmarks shows that our CCL-SLR achieves state-of-the-art performance.
Pre-training has been proven to be effective in boosting the performance of Isolated Sign Language Recognition (ISLR). 
% While existing pre-training methods efficiently leverage the compact pose modality, they suffer from inevitable information scarcity in pose data compared with raw RGB data.
% Existing pre-training methods efficiently leverage the compact pose modality.
Existing pre-training methods solely focus on the compact pose data, which eliminates background perturbation but inevitably suffers from insufficient semantic cues compared to raw RGB videos.
% However, the presence of substantial redundancy in the RGB videos results in suboptimal representation learning with only RGB modality.
Nevertheless, learning representation directly from RGB videos remains challenging due to the presence of sign-irrelevant visual features. %inherent feature redundancy.
% However, due to substantial redundant features in the RGB modality, feature extraction leads to suboptimal representation learning.
% In this work, we aim to leverage the cross-modal consistency from both RGB and pose data via the proposed Cross-modal Consistency Learning framework~(\modelname) to learn effective sign representation through self-supervised learning with two-fold contributions.
To address this dilemma, we propose a Cross-modal Consistency Learning framework~(\modelname), which leverages cross-modal consistency between both RGB and pose modalities in a self-supervised paradigm.
%to learn effective sign representation through self-supervised learning with two-fold contributions.
% It employs contrastive learning for instance discrimination within and across modalities, and introduces \textit{Motion-Preserving Masking} and \textit{Semantic Positive Mining} techniques to bridge the distribution gap between the modalities. 
First, \modelname employs contrastive learning for instance discrimination within and across modalities. 
Through single-modal and cross-modal contrastive learning, \modelname gradually aligns the feature spaces of RGB and pose modalities, thereby extracting consistency sign representation.
Second, we further introduce Motion-Preserving Masking (MPM) and Semantic Positive Mining (SPM) techniques to improve cross-modal consistency from the perspective of data augmentation and sample clustering, respectively.
%, bridging the distribution gap between modalities.
Extensive experiments on four ISLR benchmarks show that \modelname achieves impressive performance, demonstrating its effectiveness. 
The code will be released to the public.

% Pre-training has emerged as a critical strategy for enhancing Isolated Sign Language Recognition (ISLR) representation learning. Existing pre-training methods solely focus on the compact pose data, which effectively eliminate background noise but inevitably suffer from insufficient semantic cues compared to raw RGB videos. Conversely, direct representation learning from RGB videos remains challenging due to redundant sign-independent visual features. To address this dilemma, we propose a novel Cross-modal Consistency Learning (CCL-SLR) framework that leverages complementary and consistency information from both RGB and pose modalities through self-supervised pre-training. CCL-SLR employs single-modal and cross-modal contrastive learning to facilitate instance discrimination within individual modalities and across different modalities, respectively. To further explore the consistency constraints in contrastive learning, we introduce two innovative techniques from the perspective of data augmentation and sample similarity: Motion-Preserving Masking (MPM), which suppresses non-semantic regions in RGB videos while preserving motion dynamics, and Semantic Positive Mining (SPM), which identifies semantically similar samples between modalities to bridge the distribution gap. Extensive experiments on four ISLR benchmarks show that \modelname achieves impressive performance, demonstrating its effectiveness. 
% The code will be released to the public.

\end{abstract}    
\section{Introduction}
\label{sec:intro}

Sign language, which combines body gestures and facial expressions to convey semantic meaning, serves as the primary communication tool within the Deaf community. To facilitate a barrier-free understanding between sign language and spoken language, extensive research~\cite{hj,zjh,tck,sam, nla-slr,signbert,signbert+,best,stc-slr} has explored sign language understanding from various perspectives.
% Among these research directions, isolated sign language recognition (ISLR), which focuses on recognizing the meaning of sign language videos at the word level to achieve fine-grained understanding, is a fundamental task in sign language understanding.
Isolated Sign Language Recognition (ISLR) is a fundamental task in sign language understanding, that focuses on recognizing the meaning of sign language videos at the word level to achieve fine-grained understanding.
For the deployment of ISLR models in real-world applications, developing robust visual encoders with enhanced representation capabilities has increasingly been recognized as a crucial research direction.

\begin{figure}[t]
    \centering
    \includegraphics[width=\linewidth]{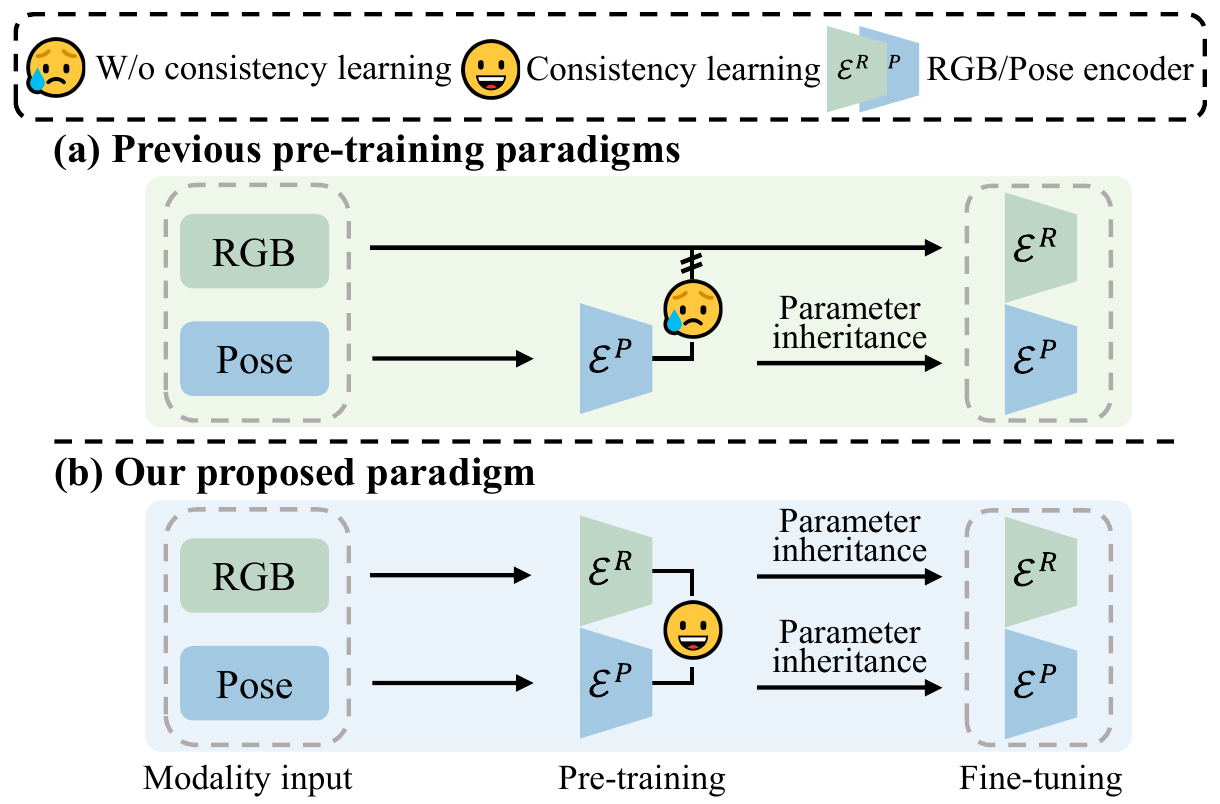}
    \caption{Comparison between previous pre-training paradigms and ours. 
    (a) Previous paradigms~\cite{best, stc-slr, signbert, signbert+} ignore modeling the interaction between RGB and pose modalities during pre-training. 
    (b) Our proposed method effectively aligns embedding spaces of two modalities via cross-modal consistency learning.}
    \label{fig:different-setup}
\end{figure}
% \input{figs/pipeline}

% Mainstream ISLR methods employ supervised paradigms, which can be categorized into two classes.
Current ISLR methods mainly employ supervised paradigms on small-scale datasets~\cite{wlasl,msasl,slr500,nmfs}, which can be categorized into two paradigms based on the input modalities.
(1)
RGB-based methods~\cite{tck,wlasl,li2020transferring,koller2018deep,msasl,hma,nmfs,li2022transcribing} explore various strategies to extract discriminative visual features from RGB videos. 
% Although intuitive, these methods fail to explicitly capture key gesture-related information. 
Although these methods are intuitive and straightforward, they fail to explicitly capture key gesture-related information. 
Furthermore, signer-dependent variations hinder effective feature extraction, ultimately compromising the robustness of the sign representations.
(2)
Pose-assisted methods~\cite{hj,zjh,sam,jiang2021skeleton,nla-slr} integrate the pose modality and employ multi-stream fusion frameworks to leverage cross-modal complementary information. 
By utilizing the compact pose modality, these methods efficiently eliminate non-skeletal noise. 
However, current pose data is directly derived from general-domain pose extractors. 
Due to motion blur and hand occlusion in sign language videos, these general-domain extractors inevitably produce errors and missing detections, limiting the performance ceiling of these pose-assisted methods.
% Specifically, NLA-SLR\cite{nla-slr} enhances sign language representation by combining pose and RGB modalities with bidirectional lateral connections.

Although these supervised methods have achieved promising results, the scarcity of labeled training data inherently constrains their representation and generalization capabilities.
%To fully leverage existing large-scale datasets, recent works~\cite{signbert,signbert+,best,stc-slr,masa} have explored self-supervised pre-training strategies in the pose modality to enhance visual modeling capabilities.
To tackle this problem, recent works~\cite{signbert,signbert+,best,stc-slr,masa} attempt to explore self-supervised pre-training strategies in the pose modality to enhance visual modeling capabilities.
For instance, MASA~\cite{masa} proposes a motion-aware masked autoencoder to utilize the pose modality by introducing a motion prior and global semantic cues into pre-training. 
% However, its effectiveness is similarly limited by the recognition accuracy of current pose estimators. 
However, it incorporates only RGB information through late fusion during fine-tuning, failing to leverage original RGB features during pre-training. 
As shown in Fig.~\ref{fig:different-setup}(a), these self-supervised methods focus solely on extracting visual information from pose modality. 
This paradigm fails to consider cross-modal feature interactions and overlooks the potential benefits of original RGB visual information.
%\textit{How to incorporate cross-modal feature interactions within the pre-training paradigm remains an unresolved challenge.}

In this paper, we aim to introduce cross-modal consistency learning into the pre-training paradigm to fully enhance feature extraction in subsequent stages.
Notably, effectively applying self-supervised paradigms such as contrastive learning to cross-modal interaction is non-trivial due to two key challenges:
(1)
% \textbf{Noise from redundant information in the RGB modality}. 
\textbf{Redundant information in the RGB modality}. 
RGB images contain redundant non-semantic information in background regions, such as signer-dependent variations such as body type and skin color, which are often irrelevant to pose information. 
%These redundant cues disrupt the cross-modal contrastive learning process, leading to the capture of ineffective consistency.
These redundant cues disrupt the extraction of sign language-related features, further impacting the ISLR performance.
(2)
% \textbf{Feature distribution inconsistency between modalities}. Although the RGB and pose modalities exhibit a one-to-one correspondence, relying solely on this alignment fails to capture equivalent intra-modal structural information due to the inherent misalignment in feature distributions across modalities.
% Addressing these two challenges is a critical step toward achieving effective cross-modal consistency learning.
% \textbf{Feature distribution inconsistency between modalities}.
\textbf{Inconsistency between modalities}.
Although the RGB and pose data of the same instance correspond to each other, the similarity of structural relationships between instances across the two modalities are inconsistent, leading to the inherent misalignment in feature distributions across RGB and pose modalities.

To this end, we propose a novel Cross-modal Consistency Learning framework (\modelname), which aims to model consistency constraints during pre-training, as shown in Fig.~\ref{fig:different-setup} (b). 
%加一段对 single-modal contrastive learning 和 cross-modal contrastive learning 的概括
We employ single-modal contrastive learning and cross-modal contrastive learning to facilitate instance discrimination within individual modalities and across different modalities, respectively.
%To effectively introduce cross-modal consistency into the contrastive learning paradigm, 
To more effectively extract consistent sign language representations from both RGB and pose modalities in the contrastive learning paradigm, our approach incorporates two key modules: Motion-Preserving Masking (MPM) and Semantic Positive Mining (SPM). 
(1)
\textbf{To eliminate impact from redundant features}, MPM introduces an off-the-shelf flow-based model~\cite{advflow} to extract latent representations, leveraging this information to capture motion dynamics. 
By preserving motion-related information, MPM suppresses non-semantic regions in RGB videos, thereby enhancing the relevance of learned features. 
(2)
\textbf{To capture cross-modal structural consistency}, SPM identifies semantically similar samples within each modality to provide pseudo-supervision for both modalities. 
Through a multi-label classification approach, this pseudo-supervision introduces shared positive constraints across different modalities, effectively bridging the distribution gap between them.
To validate the effectiveness of our method, we conduct extensive experiments on four widely adopted ISLR benchmarks. 
%and consistently achieve state-of-the-art performance.
In summary, our contributions are as follows:
\begin{itemize}
\item 
% We propose CCL-SLR, a novel pre-training framework that fully leverages single-modal and cross-modal contrastive learning on both RGB and pose modality for ISLR representation learning. 
We propose a novel pre-training framework named \modelname, which effectively leverages the single-modal and cross-modal contrastive learning on both RGB and pose modalities for enhancing ISLR.
% To the best of our knowledge, it is the \textbf{\textit{first}} pre-training framework that integrates both pose and RGB modalities.
\item We introduce MPM and SPM, two innovative techniques designed to mitigate the distribution gap between modalities and 
 further strengthen the consistency constraints during cross-modal pre-training.
\item 
% Extensive experiments on four ISLR benchmarks show that
% CCL-SLR achieves impressive performance, demonstrating
% its effectiveness.
We conduct extensive experiments on four widely-adopted benchmarks and achieve impressive performance, which demonstrates effectiveness and representation capability of \modelname.
\end{itemize}
\section{Related Work}
\label{sec:related_work}

\begin{figure*}[t]
    \centering
    \includegraphics[width=\linewidth]{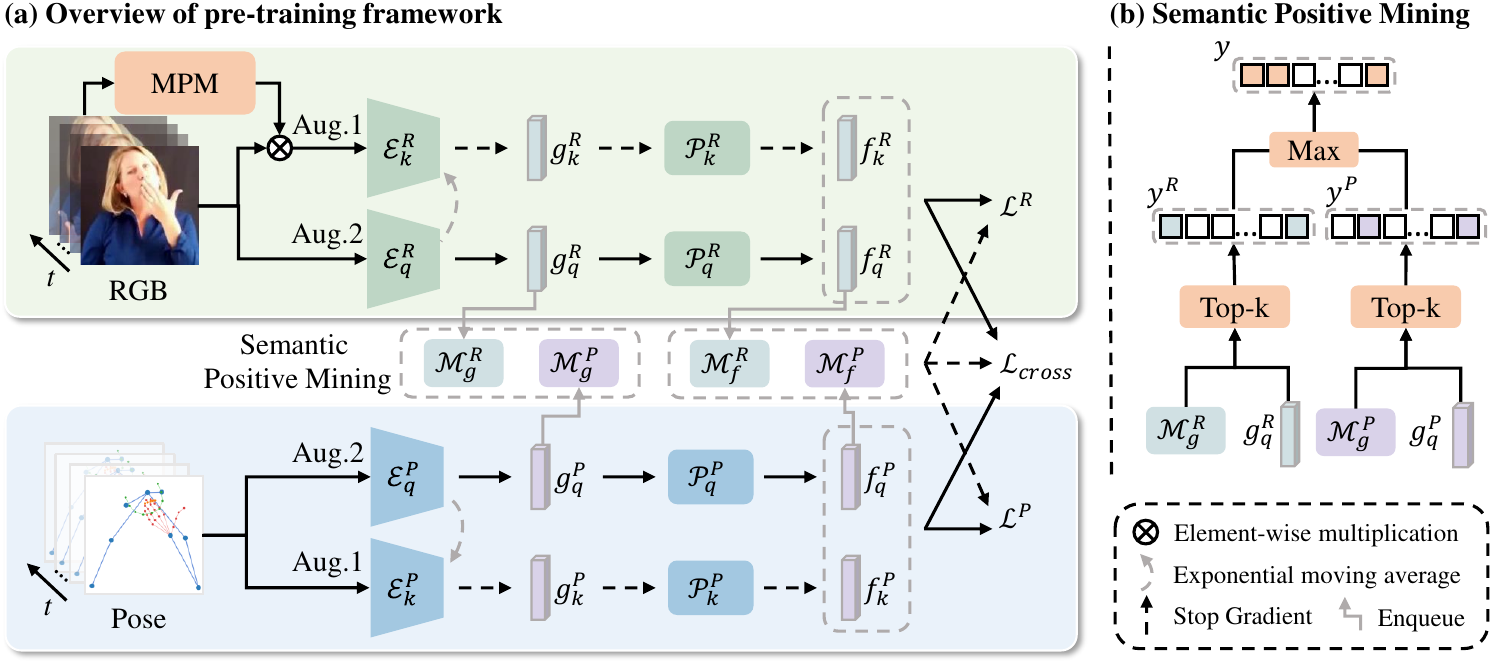}
    \caption{(a)~The CCL-SLR pre-training framework. RGB and pose sequences are processed through two single-modal branches to learn modality-specific representation. During this process, MPM is applied to mask motion-irrelevant regions in the RGB videos. Additionally, cross-modal positive samples are identified through SPM, which are utilized for single-modal loss $\mathcal{L}^R$ and $\mathcal{L}^L$. Finally, the consistency of RGB and pose embedding spaces is constrained through  cross-modal contrastive loss $\mathcal{L}_{cross}$. (b)~An overview of the Semantic Positive Mining. SPM selects the top-$k$ most similar samples from each modality-specific memory bank and merges them to get pseudo-labels.}
    % \label{different-setup}
    \label{fig:pipeline}
\end{figure*}
\noindent \textbf{Sign Language Recognition.} 
Sign language recognition is a subtask in the field of sign language understanding, which aims to recognize a single sign word within a given video. 
% Visual feature extraction plays a crucial role in ISLR. 
Early works~\cite{farhadi2007transfer,fillbrandt2003extraction,starner1995visual,ong2005automatic,rastgoo2021sign} utilize hand-crafted features to model spatio-temporal representation. With the development of deep convolution neural networks (CNNs), numerous recent works~\cite{wlasl,li2020transferring,koller2018deep,msasl,hma,nmfs,li2022transcribing} leverage CNN-based architectures such as TSM~\cite{lin2019tsm}, R3D~\cite{r3d}, and I3D~\cite{i3d} to extract visual representations from RGB videos. For example, StepNet~\cite{shen2024stepnet} uses TSM to model part-level spatial and temporal relationships, achieving impressive performance on ISLR.

However, these RGB-based methods may suffer from static irrelevant cues in sign language videos~\cite{nla-slr}, such as background and clothing. Consequently, several works have begun exploring pose-assisted methods, leveraging the compact and informative nature of sign pose data. For instance, 
%SignBERT~\cite{signbert} enhance model representation capacity through generative pretext task specifically designed for pose data during self-supervised pre-training. BEST~\cite{best} leverage discrete variational autoencoder to compress hand shapes into a codebook for implicit clustering, serving as pseudo-labels to better learn skeletal priors. Meanwhile, 
NLA-SLR~\cite{nla-slr} presents a video-keypoint multi-stream backbone and utilize semantic information within sign glosses to promote ISLR. SAM-v2~\cite{sam} jointly leverages pre-extracted multi-modal data with a branching ensemble model to achieve a higher recognition rate. Although these works effectively fuse RGB and pose modality-specific features, they rely on complex model designs and feature fusion strategies, which limits their scalability.
Meanwhile, some studies~\cite{signbert,signbert+,masa,best,stc-slr} incorporate self-supervised pre-training paradigms into ISLR to address the challenge of data scarcity.
SignBERT~\cite{signbert} enhances model representation capacity through a generative pretext task specifically designed for pose data during self-supervised pre-training. BEST~\cite{best} leverage discrete variational autoencoder to compress hand shapes into a codebook for implicit clustering, serving as pseudo-labels to better learn skeletal priors.
Nevertheless, these approaches overlook the benefits of aligning consistency visual representations of RGB and pose data during pre-training for downstream ISLR tasks.

In this work, we focus on the ISLR pre-training paradigm, learning instance discriminative representations through contrastive learning within and across modalities.

\noindent \textbf{Contrastive Representation Learning.} 
% Single-modal contrastive learning has been widely-adopt in pre-training phase on diverse image and video tasks~\cite{ssvc,multilabel,stc-slr,mocov2}. To enhance instance discriminative ability, some works~\cite{multilabel} uncovers semantic similar samples for reducing noise from false negative samples. From another perspective, some works~\cite{ssvc} enhance the representation obtained by contrastive learning through removing irrelevant static cues from the samples. 
Contrastive learning has significantly advanced self-supervised visual representation learning by performing instance discrimination in a fully self-supervised manner~\cite{he2020momentum,mocov2,simclrv2,zhang2022dino,grill2020bootstrap,radford2021learning}.
It pulls positive samples closer and pushes negative samples away.
Since there are no supervised labels, different data augmentations of the same instance are treated as a positive pair, while samples from other instances are used as negative pairs.
Recently, MoCo~\cite{he2020momentum} and MoCo v2~\cite{mocov2} have strengthened contrastive learning by maintaining a dynamic memory bank of negative sample embedding from previous iterations.
% Recent improvements in representation capability have been achieved through various strategies, such as positive mining~\cite{crossclr,multilabel} and data augmentation~\cite{ssvc}. 
% In this work, we extend the MoCo framework to explore contrastive learning for cross-modal sign language representation learning.
Building on these methods, current advances have demonstrated substantial improvements in representation capability through various technical innovations, including positive mining~\cite{crossclr,multilabel,zhang2022contrastive,jeon2021mining} and data augmentation~\cite{ssvc,ding2022motion,wang2022long,zhu2023modeling}.

Despite their success in various downstream tasks, integrating the contrastive learning paradigm into multi-modal ISLR pre-training remains challenging.
To this end, we propose \modelname, which extends the MoCo v2 framework to the challenging task of multi-modal sign language representation learning.

\section{Method}
\label{sec:method}
% As illustrated in Fig.~\ref{fig:pipeline}, the pipeline employs modality-specific data augmentation to generate augmented samples.
% \subsection{Overview.}
% \noindent \textbf{Overview.} 
As illustrated in Fig.~\ref{fig:pipeline}, our framework first applies modality-specific data augmentation to generate augmented samples.
% Subsequently, these samples are processed through their respective encoders for feature extraction.
% Next, the augmented samples then pass through their corresponding encoders for feature extraction.
Subsequently, these augmented samples are fed into their respective encoders for feature extraction.
% The outputs of encoders are utilized for memory-bank contrastive loss to learn instance discriminative representation within and across modalities.
The encoder outputs are then utilized to compute the contrastive loss, facilitating the learning of instance discriminative representations within and across modalities.
For the RGB modality, we introduce a novel Motion-Preserving Masking technique to suppress redundant non-semantic visual cues in RGB videos.
Furthermore, to enhance single-modal representation learning while leveraging cross-modal consistency, we propose a Semantic Positive Mining strategy that identifies semantically similar positive pairs.
% Further, we design a Motion-Preserving Masking (MPM) mechanism to mitigate redundant non-semantic visual cues in the RGB video.

\subsection{Preliminaries}
Since \modelname involves two modalities, we first introduce contrastive learning in the single-modal case to simplify the explanation. Single-modal contrastive learning has emerged as a powerful paradigm across various tasks, with notable success demonstrated by MoCo v2~\cite{mocov2}. 
Given a single-modal input $x$ in a mini-batch, the learning process begins with data augmentation to generate two distinct samples $x_q$ and $x_k$ (query and key). 
Two encoders, $\mathcal{E}_q$ and $\mathcal{E}_k$, are then applied to embed $x_q$ and $x_k$ into a hidden space: 
% \begin{align}
%     z_q &= \mathcal{E}_q(x_q;\theta_q), \\
%     z_k &= \mathcal{E}_k(x_k;\theta_k),
% \end{align}
\begin{equation}
    z_q = \mathcal{E}_q(x_q;\theta_q), \quad
    z_k = \mathcal{E}_k(x_k;\theta_k),
\end{equation}
where $\mathcal{E}_q$ and $\mathcal{E}_k$ denote the query and key encoders, $\theta_q$ and $\theta_k$ denote the learnable parameters of the encoders. 
Notably, the key encoder $\mathcal{E}_k$ is maintained as an exponential moving average (EMA) of the query encoder $\mathcal{E}_q$, updated as: 
% $\theta_k \gets \gamma\theta_k + (1-\gamma)\theta_q$, 
\begin{equation}
    \theta_k \gets \gamma\theta_k + (1-\gamma)\theta_q, 
\end{equation}
where $\gamma$ denotes the momentum coefficient.
During the pre-training phase, instance discrimination is achieved through the InfoNCE~\cite{infonce} loss:
\begin{equation}
\label{eq:cl_loss}
\begin{small}
\mathcal{L}_{cl}(z_q, z_k, \mathcal{S}) = 
-\log \frac{\exp(z_q {\cdot} z_k / \tau)}{\exp(z_q {\cdot} z_k / \tau) + \sum\limits_{s_i \in \mathcal{S}} \exp(z_q {\cdot} s_i / \tau)},
\end{small}
\end{equation}
where $\tau$ represents the temperature coefficient~\cite{hinton2015distilling}, and $\mathcal{S}$ denotes the negative sample embedding set maintained in a memory bank of size $N$.
The similarity between samples is computed using the dot product operation.
% \noindent \textbf{RGB Encoder.\label{sec:single_branch}} 
% We employs \text{R3D-50}~\cite{r3d} with a lightweight head network as our RGB encoders.

% \noindent \textbf{Pose Encoder.\label{sec:single_branch}} 

% \noindent \textbf{Memory Bank.\label{sec:single_branch}} 

\subsection{Single-Modal Contrastive Learning}
\label{subsec:SLF}

\noindent \textbf{Framework Architecture.\label{sec:single_branch}}
As illustrated in Fig.~\ref{fig:pipeline}, \modelname consists of two parallel single-modal branches: an RGB branch and a pose branch. 
Each branch encompasses two pathways for key and query samples, along with two queue-based memory banks. 
Each pathway consists of an encoder and a projection head. 
% Since recent ISLR  datasets do not provide pose annotation, we first extract the corresponding pose sequence $X^P \in \mathbb{R}^{T \times J \times C_p}$ using an off-the-shelf estimator~\cite{mmpose2020} from RGB frame sequence $X^R \in \mathbb{R}^{T\times H \times W \times 3}$. 
We extract the corresponding pose sequence $X^P \in \mathbb{R}^{T \times J \times C_p}$ from the RGB frame sequence $X^R \in \mathbb{R}^{T\times H \times W \times 3}$ using an off-the-shelf estimator~\cite{mmpose2020}. 
Here, \( T \) denotes the frame length, \( J\) denotes the number of joints, and \( H \) and \( W \) correspond to the spatial dimensions. $C_p=3$ represents the XY coordinates along with the confidence score of pose data.

First, we use spatial-temporal augmentation to generate different query and key samples for each single-modal branch, denoted as $X^m_s$, where $m \in \{R, P\}$ represents RGB and pose modalities and $s \in \{q,k\}$ represents query and key samples.
Second, each sample $X^m_s$ passes through its corresponding encoder $\mathcal{E}^m_s$ to obtain an embedding ${g}^m_s \in \mathbb{R}^{d}$. The RGB encoder employs \text{R3D-50}~\cite{r3d} backbone, while the pose encoder utilizes a GCN+Transformer architecture~\cite{zhou2024scaling}.
Then, ${g}^m_s$ is projected using a multi-layer perceptron (MLP) projection head $\mathcal{P}^m_s$ to obtain the final embedding $f^m_s\in \mathbb{R}^{d}$.
Two queue-based memory banks $\mathcal{M}^m_g \in \mathbb{R}^{N \times d}$ and $\mathcal{M}^m_f \in \mathbb{R}^{N \times d}$ are utilized for storing L2-normalized embeddings of ${g}^m_k$ and ${f}^m_k$, respectively:
\begin{equation}
    \Vert{g}^m_k\Vert_2 \overset{Enqueue}\Longrightarrow \mathcal{M}^m_g, \quad
    \Vert{f}^m_k\Vert_2 \overset{Enqueue}\Longrightarrow \mathcal{M}^m_f, \label{banks}
\end{equation}
where $\Vert\cdot\Vert_2$ denotes the L2 normalization operation. 
Here, $\mathcal{M}^m_g$ is utilized to mine semantic positive samples, and $\mathcal{M}^m_f$ is applied for contrastive learning.
% To perform intra-modal instance discrimination, we formulate the single-modal contrastive loss for both modalities as follows:
% \begin{align}
%     \mathcal{L}^{R}_{cl} &= \mathcal{L}_{cl}({f}^R_q, {f}^R_k, \mathcal{M}^R_f), \\
%     \mathcal{L}^{P}_{cl} &= \mathcal{L}_{cl}({f}^P_q, {f}^P_k, \mathcal{M}^P_f).
% \end{align}

\begin{figure}[t]
    \centering
    \includegraphics[width=\linewidth]{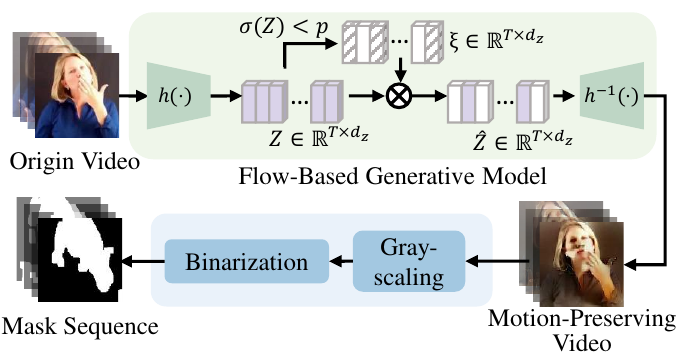}
    \caption{An overview of MPM. $h(\cdot)$ and $h^{-1}(\cdot)$ denote the transformation from pixel/latent space to latent/pixel space, respectively. The video is encoded into latent space using a flow-based generative model. Channels in the latent representation with low standard deviation along the temporal dimension are then masked out. The masked latent representation is decoded back into pixel space. The reconstructed frames are then processed through gray-scaling and binarization to produce motion-aware mask sequence.}
    \label{fig:mmaug}
    %\vspace{-10pt}
\end{figure}

\noindent \textbf{Motion-Preserving Masking (MPM).\label{sec:smg}} 
Unlike pose sequences, which focus on keypoint coordinates, RGB videos contain substantial non-gestural information such as background and clothing.
To preserve essential motion components while suppressing irrelevant regions, we propose MPM, a data augmentation technique that enhances motion capture capabilities and mitigates bias from non-skeletal elements.
% MPM employs a pre-trained flow-based generative model~\cite{advflow} for motion-preserving latent representation extraction.
As shown in Fig.\ref{fig:mmaug}, MPM employs a pre-trained flow-based generative model, AdvFlow~\cite{advflow}, to transform original RGB videos into motion-preserving videos and generate a mask sequence for data augmentation.
% For mapping RGB videos from pixel space to a compact feature space, MPM employs a pre-trained flow-based generative model, AdvFlow~\cite{advflow} for motion-preserving latent representation extraction.

% Concretely, given an RGB video $X^R$, we encode it into latent representation $Z \in \mathbb{R}^{T \times d_z}$ through:
% \begin{equation}
%     Z = h(X^R),
% \end{equation}
% where $h(\cdot)$ denotes the transformation from video space to hidden space.
Concretely, given an RGB video $X^R$, we encode it into a latent representation $Z \in \mathbb{R}^{T \times d_z}$.
Empirically, the temporal standard deviation of each channel reliably indicates its motion content. 
We leverage this measure on $Z$ to preserve dynamic features while suppressing static ones:
% \begin{align}
%     \xi &= \mathds{1}(\sigma(Z) < p), \\
%     \hat{Z} &= Z \otimes \xi,
% \end{align}
\begin{equation}
    \hat{Z} = Z \otimes \mathds{1}(\sigma(Z) < p),
\end{equation}
where $\mathds{1}(\cdot)$ denotes the indicator function, $\otimes$ denotes the element-wise multiplication, $\sigma(\cdot)$ computes the standard deviation along the temporal channel, and $p$ controls the mask intensity. 
% $\hat{Z} \in \mathbb{R}^{T \times d_z}$ is masked hidden representation. 
% The motion-preserving video $\hat{X}^R \in \mathbb{R}^{T \times H \times W \times 3}$ is then decoded from hidden space as follows:
% \begin{equation}
%     \hat{X}^R = h^{-1}(\hat{Z}),
% \end{equation}
% where $h^{-1}(\cdot)$ represents the transformation from hidden space to video space.
% The motion-preserving video $\hat{X}^R \in \mathbb{R}^{T \times H \times W \times 3}$ is subsequently decoded from the masked hidden representation $\hat{Z}$.
% The final binary mask sequence $\zeta \in \mathbb{R}^{T \times H \times W \times 1}$ is generated by applying gray-scaling and binarization to $\hat{X}^R$.
% The final augmented sample is obtained by applying $\zeta$ to mask the original RGB video $X^R$ with probability $\alpha$, where $\alpha$ is a hyperparameter.
The motion-preserving video $\hat{X}^R \in \mathbb{R}^{T \times H \times W \times 3}$ is then decoded from the masked latent representation $\hat{Z}$. 
Subsequently, we derive the binary mask sequence $\zeta \in \mathbb{R}^{T \times H \times W}$ from $\hat{X}^R$ and apply it to $X^R$ with probability $\alpha$ to generate the augmented sample.

\noindent \textbf{Semantic Positive Mining (SPM).\label{sec:spm}} 
While single-modal contrastive learning enables \modelname to extract modality-specific representations, the ``semantic collision'' problem identified by CrosSCLR~\cite{crossclr} remains a challenge.
The semantic collision occurs during contrastive learning when sample embeddings with high semantic similarity in the memory bank are mistakenly treated as negative examples.
To mitigate this semantic misalignment, we introduce SPM, a mechanism that identifies semantic positive samples across memory banks of different modalities.
As illustrated in Fig.~\ref{fig:pipeline} (b), we select the top-$k$ nearest neighbors of ${g}^m_q$ from $\mathcal{M}^m_g$ as positive samples, while treating the remaining embeddings as negative samples.
This selection process generates binary pseudo-labels $y^m \in \mathbb{R}^{N}$ for both modalities:
\begin{equation}
    y^R = Top\text{-}k({g}^R_q \cdot \mathcal{M}^R_g),\quad
    y^P = Top\text{-}k({g}^P_q \cdot \mathcal{M}^P_g).
\end{equation}
% \begin{equation}
%     y^m = Top\text{-}k({g}^m_q \cdot \mathcal{M}^m_g).
% \end{equation}
% To fully leverage multi-modal consistency constraints, we combine  $y^R$ and $y^P$ through maximization to obtain the cross-modal binary pseudo-label: 
To align the neighborhood structure across modalities, we combine  $y^R$ and $y^P$ by taking their element-wise maximum to obtain the cross-modal binary pseudo-label $y$.
% \begin{equation}
%     y = \max(y^R,\ y^P).
%     % y = y^R \odot y^P.
% \end{equation}
% Actually, since the labels are binary, the max function here is equivalent to taking the union of the positive samples from two modalities.
The embeddings in $\mathcal{M}^m_f$ serve as classifiers, generating logits $\hat{y}^m \in \mathbb{R}^{N}$ that classify ${f}^m_q$ into \(N\) pseudo categories:
\begin{equation}
    \hat{y}^m = \phi(f^m_q \cdot \mathcal{M}_f^m / \tau_{spm}),
\end{equation}
where $\tau_{spm}$ denotes the temperature coefficient and $\phi(\cdot)$ is the sigmoid function~\cite{zhu2017learning}.
Then, $\hat{y}^m$ and $y$ are utilized for binary cross-entropy (BCE) loss to help single-modal contrastive learning bring the query sample closer to semantic positive samples from both modalities:
\begin{equation}
\begin{small}
    \mathcal{L}^m_{spm} = -\frac{1}{N}\sum_{i=1}^{N} \left[y_i \log \hat{y}^m_i + (1 - y_i) \log(1 - \hat{y}^m_i)\right],
\end{small}
\end{equation}
Additionally, to perform intra-modal instance discrimination, we formulate the single-modal contrastive loss for both modalities as follows:
\begin{equation}
\mathcal{L}^{m}_{cl} = \mathcal{L}_{cl}({f}^m_q, {f}^m_k, \mathcal{M}^m_f).
    % \mathcal{L}^{R}_{cl} &= \mathcal{L}_{cl}({f}^R_q, {f}^R_k, \mathcal{M}^R_f), \\
    % \mathcal{L}^{P}_{cl} &= \mathcal{L}_{cl}({f}^P_q, {f}^P_k, \mathcal{M}^P_f).
\end{equation}
The final single-modal loss is formulated as::
\begin{equation}
    \mathcal{L}_{single} = \underbrace{\mathcal{L}^R_{cl} + \mathcal{L}^R_{spm}}_{\displaystyle \mathcal{L}^R} + \underbrace{\mathcal{L}^P_{cl} + \mathcal{L}^P_{spm}}_{\displaystyle \mathcal{L}^P}.
\end{equation}

\subsection{Cross-Modal Contrastive Learning}
% Multi-modal data inherently contains rich complementary information, which is essential to learn modality-invariant representation during pre-training.
While previous work \cite{zhao2024self} successfully employed Kullback-Leibler (KL) divergence to ensure distribution consistency within the pose modality, cross-modal alignment between pose and RGB modality remains challenging due to the substantial distribution gap between them.
To this end, we extend single-modal contrastive learning to the cross-modal domain by directly aligning embeddings from both modalities in the latent space.
Specifically, for RGB features $f^R_q$, we utilize $f^P_k$ as positive samples while treating embeddings in $\mathcal{M}^P_f$ as negative.
A symmetric procedure is applied to pose features.
The bidirectional cross-modal contrastive loss functions are defined as:
\begin{align}
    \mathcal{L}^{R\rightarrow P} &= \mathcal{L}_{cl}(f^P_q, f^R_k, \mathcal{M}^R_f), \\
    \mathcal{L}^{P\rightarrow R} &= \mathcal{L}_{cl}(f^R_q, f^P_k, \mathcal{M}^P_f), \\
    \mathcal{L}_{cross} &= \mathcal{L}^{R\rightarrow P} + \mathcal{L}^{P\rightarrow R}. 
\end{align} 
The overall pre-training objective function is formulated as:
\begin{equation}
    \mathcal{L} = \mathcal{L}_{single} + \mathcal{L}_{cross}.
\end{equation}
% This pre-training strategy enables \modelname to align embedding spaces within and across modalities, thereby performing instance discrimination for enhanced consistency representation.
This pre-training strategy enables \modelname to align embedding spaces across modalities, performing instance discrimination for consistency representation learning.

% \subsection{Fine-Tuning}
% \label{subsec:FT}
% During fine-tuning phase, we retain only the query encoders as feature extractors and replace the MLP layers with fully connected layers.
% Both branches are optimized with cross-entropy loss.
% The final prediction is obtained by summing the predicted logits from the RGB and pose branches.
\subsection{Model Details}
\label{subsec:FT}
During the pre-training phase, each projection head is a 2-layer MLP which projects the encoder outputs into a 128-dimensional embedding. 
During the fine-tuning phase, we employ the query encoders as feature extractors and replace the projection head with a fully connected layer for prediction.
In this phase, both branches are optimized via cross-entropy loss. 
The final prediction is obtained by summing the predicted logits from the RGB and pose branches.

\section{Experiments}

\begin{table*}[t]
% \caption{Comparison with previous works on WLASL and MSASL. The results of I3D and ST-GCN are reproduced by~\cite{} and~\cite{}, respectively.(``$\dagger$'' denotes methods using extra data. ``$*$'' denotes methods using more modalities, \eg, optical flow, depth map and depth flow. ``$\ddagger$'' indicates the model with self-superwised learning.)}
\centering
\resizebox{\linewidth}{!}{
\begin{tabular}{l|cc|cc|cc|cc|cc|cc|cc|cc|cc}
\toprule[1.0pt]
\multirow{3}{*}{Method} & \multicolumn{2}{c|}{Modality} &\multicolumn{4}{c|}{WLASL2000} & \multicolumn{4}{c|}{WLASL300} & \multicolumn{4}{c|}{MSASL1000} & \multicolumn{4}{c}{MSASL200} \\
\cmidrule{2-19}
& \multirow{2}{*}{\hspace{-0.3em}Pose\hspace{-0.5em}}& \multirow{2}{*}{\hspace{-0.5em}RGB\hspace{-0.3em}} & \multicolumn{2}{c|}{Per-instance} & \multicolumn{2}{c|}{Per-class} & \multicolumn{2}{c|}{Per-instance} & \multicolumn{2}{c|}{Per-class} & \multicolumn{2}{c|}{Per-instance} & \multicolumn{2}{c|}{Per-class} & \multicolumn{2}{c|}{Per-instance} & \multicolumn{2}{c}{Per-class} \\
& & & Top-1 & Top-5 & Top-1 & Top-5 & Top-1 & Top-5 & Top-1 & Top-5 & Top-1 & Top-5 & Top-1 & Top-5 & Top-1 & Top-5 & Top-1 & Top-5 \\
\midrule
ST-GCN~\cite{stgcn} & \Pmodal & 34.40 & 66.57 & 32.53 & 65.45 & 44.46 & 73.05 & 45.29 & 73.16 & 36.03 & 59.92 & 32.32 & 57.15 & 52.91 & 76.67 & 54.20 & 77.62 \\
I3D~\cite{i3d} & \Rmodal & 32.48 & 57.31 & - & - & 56.14 & 79.94 & - & - & - & - & 57.69 & 81.08 & - & - & 81.97 & 93.79 \\
TCK~\cite{tck} & \Rmodal & - & - & - & - & 68.56 & 89.52 & 68.75 & 89.41 & - & - & - & - & 80.31 & 91.82 & 81.14 & 92.24 \\
StepNet~\cite{shen2024stepnet} & \Rmodal & 56.89 & 88.64 & 54.54 & 87.97 & 74.70 & 91.02 &  75.32 & 91.17 & - & - & - & - & - & - & - & - \\
StepNet$^\ddag$~\cite{shen2024stepnet} & \Rmodal & 61.17 & 91.94 & 58.43 & 91.43 & - & - &  - & - & - & - & - & - & - & - & - & - \\
% ST-GCN~\cite{stgcn} & \Pmodal & 34.40 & 66.57 & 32.53 & 65.45 & 44.46 & 73.05 & 45.29 & 73.16 & 36.03 & 59.92 & 32.32 & 57.15 & 52.91 & 76.67 & 54.20 & 77.62 \\
HMA~\cite{hma} & \RPmodal & 51.39 & 86.34 & 48.75 & 85.74 & - & - & - & - & 69.39 & 87.42 & 66.54 & 86.56 & 85.21 & 94.41 & 86.09 & 94.42 \\
% TCK~\cite{tck} & \Rmodal & - & - & - & - & 68.56 & 89.52 & 68.75 & 89.41 & - & - & - & - & 80.31 & 91.82 & 81.14 & 92.24 \\
BEST~\cite{best} & \RPmodal & 54.59 & 88.08 & 52.12 & 87.28 & 75.60 & 92.81 & 76.12 & 93.07 & 71.21 & 88.85 & 68.24 & 87.98 & 86.83 & 95.66 & 87.45 & 95.72 \\
SignBERT~\cite{signbert} & \RPmodal & 54.69 & 87.49 & 52.08 & 86.93 & 74.40 & 91.32 & 75.27 & 91.72 & 71.24 & 89.12 & 67.96 & 88.40 & 86.98 & 96.39 & 87.62 & 96.43 \\
SignBERT+~\cite{signbert+} & \RPmodal & 55.59 & 89.37 & 53.33 & 88.82 & 78.44 & 94.31 & 79.12 & 94.43 & 73.71 & 90.12 & 70.77 & 89.30 & 88.08 & 96.47 & 88.62 & 96.47 \\
SAM$^*$($5^\dag$)~\cite{jiang2021skeleton} & \RPmodal & 58.73 & 91.46 & 55.93 & 90.94 & - & - & - & - & - & - & - & - & - & - & - & - \\
SAM-v2$^*$($5^\dag$)~\cite{sam} & \RPmodal & 59.39 & 91.48 & 56.63 & 90.89 & - & - & - & - & - & - & - & - & - & - & - & - \\
NLA-SLR~\cite{nla-slr} & \RPmodal & 61.05 & 91.45 & 58.05 & 90.70 & 86.23 & 97.60 & 86.67 & 97.81 & 72.56 & 89.12 & 69.86 & 88.48 & 88.74 & 96.17 & 89.23 & 96.38 \\
NLA-SLR($3^\dag$)~\cite{nla-slr} & \RPmodal & 61.26 & 91.77 & 58.31 & 90.91 & 86.98 & 97.60 & 87.33 & 97.81 & 73.80 & 89.65 & 70.95 &  89.07 & 89.48 & 96.69 & 89.86 & 96.93 \\
\midrule
CCL-SLR (Ours) & \RPmodal & \textbf{62.20} & \textbf{92.39} & \textbf{58.80} & \textbf{91.85} & \textbf{86.28} & \textbf{97.63} & \textbf{86.81} & \textbf{97.87} & \textbf{77.71} & \textbf{92.28} & \textbf{76.27} & \textbf{92.30} & \textbf{90.21} & \textbf{96.47} & \textbf{90.86} & \textbf{96.65} \\
\bottomrule[1.0pt]
\end{tabular}
}
\caption{Comparison with previous works on WLASL and MSASL datasets. The results of ST-GCN and I3D are reproduced by \cite{signbert} and \cite{wlasl}, respectively. $*$ denotes methods using extra modalities such as optical flow, depth map and depth flow. $\dag$ denotes a multi-crop inference. $\ddag$ denotes fusion results with optical flow.
%The results of ST-GCN and I3D are reproduced by~\cite{signbert}  and~\cite{msasl}.
}
\label{tab:wlasl-msasl}
%\vspace{-5pt}
\end{table*}
% % NMFs SLR500
\begin{table}[t]
\centering
\resizebox{\linewidth}{!}{
\begin{tabular}{l | cc| cc | c}
\toprule[1.0pt]
\multirow{2}{*}{Method} & \multicolumn{2}{c|}{Modality} & \multicolumn{2}{c|}{NMFs-CSL} & {SLR500} \\
\cmidrule{2-6}
 & Pose & RGB & Top-1 & Top-5 & Top-1  \\
\midrule
ST-GCN$^*$~\cite{stgcn} & \Pmodal& 59.9 & 86.8 & 90.0  \\
I3D$^*$~\cite{i3d} & \Rmodal & 64.4 & 88.0 & -  \\
GLE-Net~\cite{nmfs} & \Rmodal& 69.0 & 88.1 & 96.8  \\
StepNet~\cite{shen2024stepnet} & \Rmodal& 77.2 & 92.5 & - \\
StepNet$^\ddag$~\cite{shen2024stepnet} & \Rmodal& 83.6 & 97.0 & - \\
HMA~\cite{hma} & \RPmodal& 75.6 & 95.3 & 95.9  \\
SignBERT~\cite{signbert} & \RPmodal& 78.4 & 97.3 & 97.6  \\
SignBERT+~\cite{signbert+} & \RPmodal& - & - & 97.8  \\
BEST~\cite{best} & \RPmodal& 79.2 & 97.1 & 97.7  \\
NLA-SLR~\cite{nla-slr} & \RPmodal& 83.4 & 98.3 & - \\
NLA-SLR(3$^\dag$)~\cite{nla-slr} & \RPmodal& 83.7 & 98.5 & - \\
\midrule
CCL-SLR (Ours) \hspace{1em} & \RPmodal & \textbf{84.4} & \textbf{99.3} & \textbf{97.8} \\
\bottomrule[1.0pt]
\end{tabular}
}
\caption{Comparison on NMFs-CSL and SLR500 datasets. $\dag$ denotes a multi-crop inference. $\ddag$ denotes fusion results with optical flow. $*$ represents methods reproduced by ~\cite{nmfs}.}
\label{nmfs-slr500}
%\vspace{-10pt}
\end{table}
\label{sec:exp}
\subsection{Datasets and Evaluation Metrics}
\label{subsec:DEM}
% We evaluate our method on four public sign language datasets, MSASL~\cite{msasl}, WLASL~\cite{wlasl}, NMFs-CSL~\cite{nmfs} and SLR500~\cite{slr500}.
We evaluate our method on four public sign language datasets: MSASL~\cite{msasl}, WLASL~\cite{wlasl}, NMFs-CSL~\cite{nmfs}, and SLR500~\cite{slr500}, utilizing the entire training sets of all four datasets to pre-train our framework.

% The American Sign Language (ASL) datasets include: (1) \textit{MSASL}, consisting of 25,513 samples over 2,000 distinct sign words (16,054/5,287/4,172 for train/dev/test sets) and a subset of top 200 frequent glosses; (2) \textit{WLASL}, comprising 21,083 samples across 1,000 glosses (14,289/3,916/2,878 for train/dev/test) and a subset of 300 frequent glosses.

% The Chinese Sign Language (CSL) datasets include: (1) \textit{NMFs-CSL}, a challenging dataset incorporating fine-grained non-manual features (NMFs) comprising of 1,067 words with 32,010 samples (25,608/6,402 for train/test); (2) \textit{SLR500}, a large-scale dataset covering 500 common words with 125,000 total samples (90,000/35,000 for train/test).

% Following~\cite{signbert, nla-slr}, we report the average accuracy over instances and classes, respectively. Since NMFs-CSL and SLR500 have a balanced distribution with an equal number of samples per class, we only report per-instance accuracy.

% We leverage the entire training sets of all four datasets to pre-train our framework. 

MSASL is an American Sign Language (ASL) dataset, consisting of 25,513 samples with a vocabulary size of 1,000. It includes 16,054, 5,287, and 4,172 samples in the training, development (dev), and test set, respectively. Additionally, it provides a subset containing only the top 200 most frequent glosses. WLASL is another widely used ASL dataset with a vocabulary size of 2,000. It includes 14,289, 3,916, and 2,878 samples in the training, dev, and test sets, respectively. Similar to MSASL, WLASL also provides a subset of 300 frequent glosses. 

NMFs-CSL is a challenging Chinese Sign Language (CSL) dataset that incorporates a wide range of fine-grained non-manual features (NMFs). With a vocabulary size of 1,067, it contains 25,608 training samples and 6,402 test samples. SLR500 is another large-scale CSL dataset covering 500 common words. It contains a total of 125,000 samples, with 90,000 samples for training and 35,000 samples for testing. Following~\cite{signbert, nla-slr}, we report the average accuracy over instances and classes, respectively. Since NMFs-CSL and SLR500 exhibit a balanced distribution with an equal number of samples per class, we only report per-instance accuracy.

\subsection{Implementation Details}
\label{subsec:ES}
% In each frame, the pose input includes 77 keypoints, which is consistency with CoSign \cite{cosign}. 
% Besides, all RGB frames are resized to $224 \times 224$. 
% In pre-training phase, we sample a fixed number of frames $T = 32$ from each origin sequence. 
% Spatial augmentation within the RGB branch encompasses random cropping, random horizontal flipping, color jittering. 
% Similarly, the spatial augmentation of the Pose branch consists of a random combination of rotation, scaling, joint masking, and random flipping. 
% The temporal augmentation for both branches involves random temporal cropping. 

% The temperature coefficients $\tau$/$\tau_{spm}$ are 0.07/0.02, respectively. 
% The hyper-parameters $\alpha$/$p$ of MMAug are 0.5/0.2, respectively. 
% During pre-training, we employ the SGD optimizer with a momentum of 0.9. 
% The initial learning rate is 0.01 with batch size 64, and divided by 10 after 100 epochs. 
% The pre-training phase includes total 140 epochs.
% During fine-tuning, the SGD optimizer with the learning rate of 0.05 and 0.9 momentum is applied. 
% The learning rate is reduced by a factor of 0.1 after 25/35 epochs. 
% The batch size is set to 32 with total 40 epochs.
% All experiments are performed with NVIDIA RTX 3090 by PyTorch.
% For data preprocessing, we select 75 keypoints for pose input per frame and resize all RGB frames to $224 \times 224$. 
The RGB branch applies spatial augmentations including random cropping, horizontal flipping, and color jittering, while the pose branch implements random rotation, scaling, joint masking, and flipping. Both branches share temporal augmentation through random temporal cropping.
%to simulate various signing speeds from different signers.
More details about data preprocessing, the pose encoder and data augmentation are provided in the Appendix.
%Appendix~\uppercase\expandafter{\romannumeral1}-A.

% The hyperparameters are configured as follows: temperature coefficients $\tau = 0.07$ and $\tau_{spm} = 0.02$; MPM parameters $\alpha = 0.5$ and $p = 0.2$.
The temperature hyperparameters $\tau$ and $\tau_{spm}$ are set to $0.07$ and $0.02$, respectively. 
The hyperparameters of MPM $p$ and $\alpha$ are set to $0.5$ and $0.2$, respectively.
During the pre-training phase, \modelname is trained for 140 epochs with a learning rate of 0.01 and batch size 64, with the learning rate reduced by a factor of 10 at epoch 100.
During the fine-tuning phase, the model is trained for 40 epochs with a learning rate of 0.05 and a batch size of 128, with the learning rate reduced by a factor of 10 at epochs 25 and 35.
A Stochastic Gradient Descent (SGD) optimizer with a momentum of 0.9 is utilized throughout the training process.
All experiments are conducted on 8$\times$ NVIDIA RTX 3090 GPUs.
% using PyTorch
% During pre-training, we set the initial learning rate to 0.01 with batch size 64, decrease it by a factor of 10 after 100 epochs, and train for 140 epochs in total. For fine-tuning, we use learning rate 0.05 with batch size 128, apply learning rate decay by 0.1 at epochs 25 and 35, and train for 40 epochs. All experiments are performed with NVIDIA RTX 3090 by PyTorch.

\subsection{Comparison with State-of-the-art Methods}
\noindent \textbf{MSASL.\label{comp:msasl}} Table~\ref{tab:wlasl-msasl} presents an extensive comparison of our method against other approaches on two MSASL sub-splits. Benefiting from the consistency modeling between dual modalities during pre-training, our method surpasses the previously best-performing NLA-SLR~\cite{nla-slr}, which utilized semantic information within sign glosses, by 5.15\%/1.53\% on the 1,000/200 sub-splits, respectively, in terms of per-instance top-1 accuracy.

\noindent \textbf{WLASL.\label{comp:wlasl}} In Table~\ref{tab:wlasl-msasl}, we also illustrates the results on the WLASL dataset. our CCL-SLR achieves a 2.81\% improvement in the accuracy of the top-1 per instance compared to SAM-v2~\cite{sam}, a multi-modal ensemble approach using RGB, pose, optical flow, and depth, demonstrating that alignment of the cross-modal distribution during pre-training enhances the generalization capability.

\noindent \textbf{NMFs-CSL.\label{comp:nmfs}} The comparison on the NMFs-CSL dataset is shown in Table~\ref{nmfs-slr500}. Unlike previous pre-training methods such as SignBERT\cite{signbert} or BEST\cite{best}, which simply fuse RGB predictions in downstream tasks and lack consideration of modeling cross-modal consistency during pre-training. Our method achieves superior performance with 6.0\% and 5.2\% improvements over SignBERT and BEST on the NMFs-CSL dataset, respectively. 

\noindent \textbf{SLR500.\label{comp:slr500}} Lastly, Table~\ref{nmfs-slr500} also presents the performance comparison on the SLR500 dataset. On this sufficiently large dataset, self-supervised learning methods such as SignBERT+\cite{signbert+} and BEST\cite{best} achieve more outstanding performance. Similarly, our cross-modal pre-training approach effectively integrates consistency cues from both modalities, achieving a top-1 accuracy of 97.8\%.

\begin{table}[!t]
\centering
\resizebox{\linewidth}{!}{
\begin{tabular}{ccc | cc | cc}
\toprule[1.0pt]
 \multirow{2}{*}{$\mathcal{L}^{m}_{spm}$} & \multirow{2}{*}{$\mathcal{L}_{cross}$} & \multirow{2}{*}{MPM} & \multicolumn{2}{c|}{Per-instance} & \multicolumn{2}{c}{Per-class}\\
 & & & Top-1 & Top-5 & Top-1 & Top-5  \\
\midrule
& & & 70.04 & 88.43 & 67.73 & 86.91  \\
\ding{52} & & & 71.20 & 88.93 & 68.96 & 88.14 \\
% \ding{52} &  & \ding{52} & 72.69 & 90.23 & 79.81 & 89.46 \\
\ding{52} & \ding{52} & & 76.27 & 91.78 & 74.26 & 91.57 \\
\ding{52} & \ding{52} & \ding{52} & \textbf{77.71} & \textbf{92.28} & \textbf{76.27} & \textbf{92.30} \\
\bottomrule[1.0pt]
\end{tabular}
}
\caption{Ablation studies on the major components of CCL-SLR.}
\label{tab:ab-main}
%\vspace{-1.5em}
\end{table}
\begin{table}[!t]
\centering
\begin{tabular}{l | cc | cc}
\toprule[1.0pt]
  \multirow{2}{*}{Setting} & \multicolumn{2}{c|}{Per-instance} & \multicolumn{2}{c}{Per-class}\\
  & Top-1 & Top-5 & Top-1 & Top-5  \\
\midrule
 w/o $\mathcal{L}_{cross}$ & 72.69 & 90.23 & 79.81 & 89.46  \\
 w/ $\mathcal{L}_{cross}$ & \textbf{77.71} & \textbf{92.28} & \textbf{76.27} & \textbf{92.30} \\
\bottomrule[1.0pt] 
\end{tabular}
\caption{Ablation studies on cross-modal contrastive learning loss.}
\label{tab:ab-ccl}
%\vspace{-1.5em}
%\vspace{-15pt}
\end{table} 

\subsection{Ablation Study}
In this section, we conduct several ablation studies to verify the effectiveness of key components in our proposed framework on the MSASL~\cite{stc-slr} dataset.

\noindent \textbf{Key Components of CCL-SLR.}
As shown in Table~\ref{tab:ab-main}, we conduct ablation studies on three key components of \modelname: Semantic Positive Mining ($\mathcal{L}^m_{spm}$), cross-modal contrastive learning ($\mathcal{L}_{cross}$), and Motion-Preserving Masking (MPM). 
Initially, with a baseline comprising two independent single-modal learning branches, the incorporation of SPM improves top-1 accuracy by 0.85\%, which is attributed to semantic positive pairs enforcing cross-modal consistency between sample spaces.
The implementation of $\mathcal{L}_{cross}$ further boosts the performance significantly from 74.66\% to 76.27\%, demonstrating its effectiveness in aligning multi-modal representation during pre-training. 
Ultimately, by suppressing non-semantic features in RGB videos, the integration of MPM achieves the optimal performance with 77.71\% top-1 accuracy.

\begin{figure}[t]
    \centering
    \includegraphics[width=\linewidth]{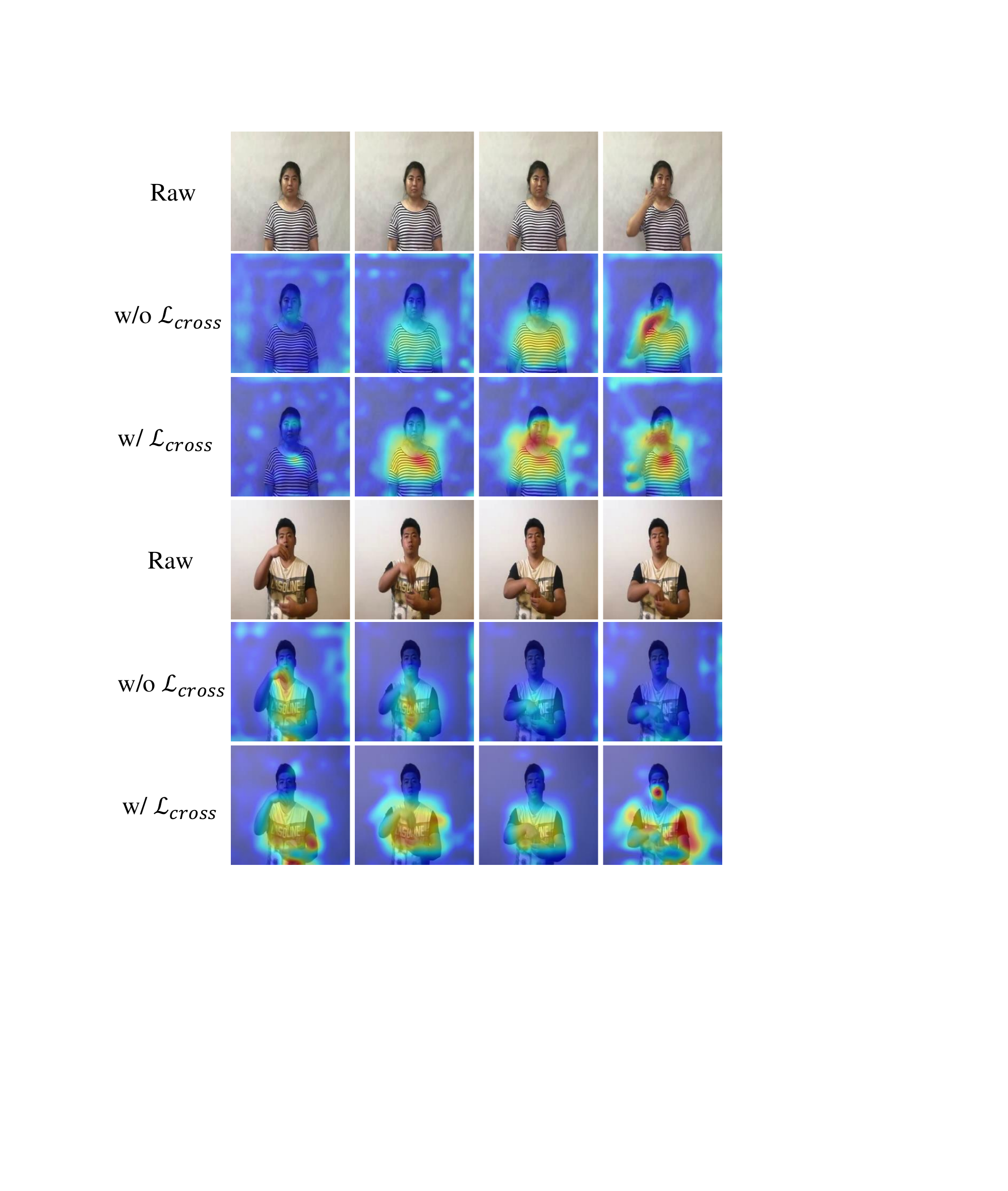}
    \caption{Visualizations of heatmaps by Grad-CAM~\cite{8237336}. Top: raw frames; Middle: heatmaps of the fine-tuning results of the RGB encoder from pre-training without $\mathcal{L}_{cross}$; Bottom: heatmaps of the fine-tuning results of the RGB encoder from pre-training with $\mathcal{L}_{cross}$. All samples are from the NMFs-CSL~\cite{nmfs} dataset.
}
\label{fig:cam}
%\vspace{-15pt}
\end{figure}

\noindent \textbf{Cross-Modal Contrastive Learning Loss.}
As shown in Table~\ref{tab:ab-ccl}, we first investigate the impact of the cross-modal contrastive learning loss $\mathcal{L}_{cross}$. After removing $\mathcal{L}_{cross}$ from CCL-SLR, the top-1 accuracy significantly drops by about 5\%. The reason is that $\mathcal{L}_{cross}$ plays a crucial role in our framework by explicitly and effectively aligning the feature spaces between the two modalities, which implicitly helps the model extract cross-modal consistency cues. Additionally, Fig.~\ref{fig:cam} demonstrates the heatmaps generated by fine-tuning RGB encoder from pre-training with and without $\mathcal{L}_{cross}$. Especially, performing cross-modal contrastive learning during pretraining allows the RGB branch to learn semantic information from the pose modality. This significantly enhances the model's attention to non-manual features (yellow and red areas) such as facial expressions and body movements.

\begin{figure}[t]
    \centering
    \includegraphics[width=\linewidth]{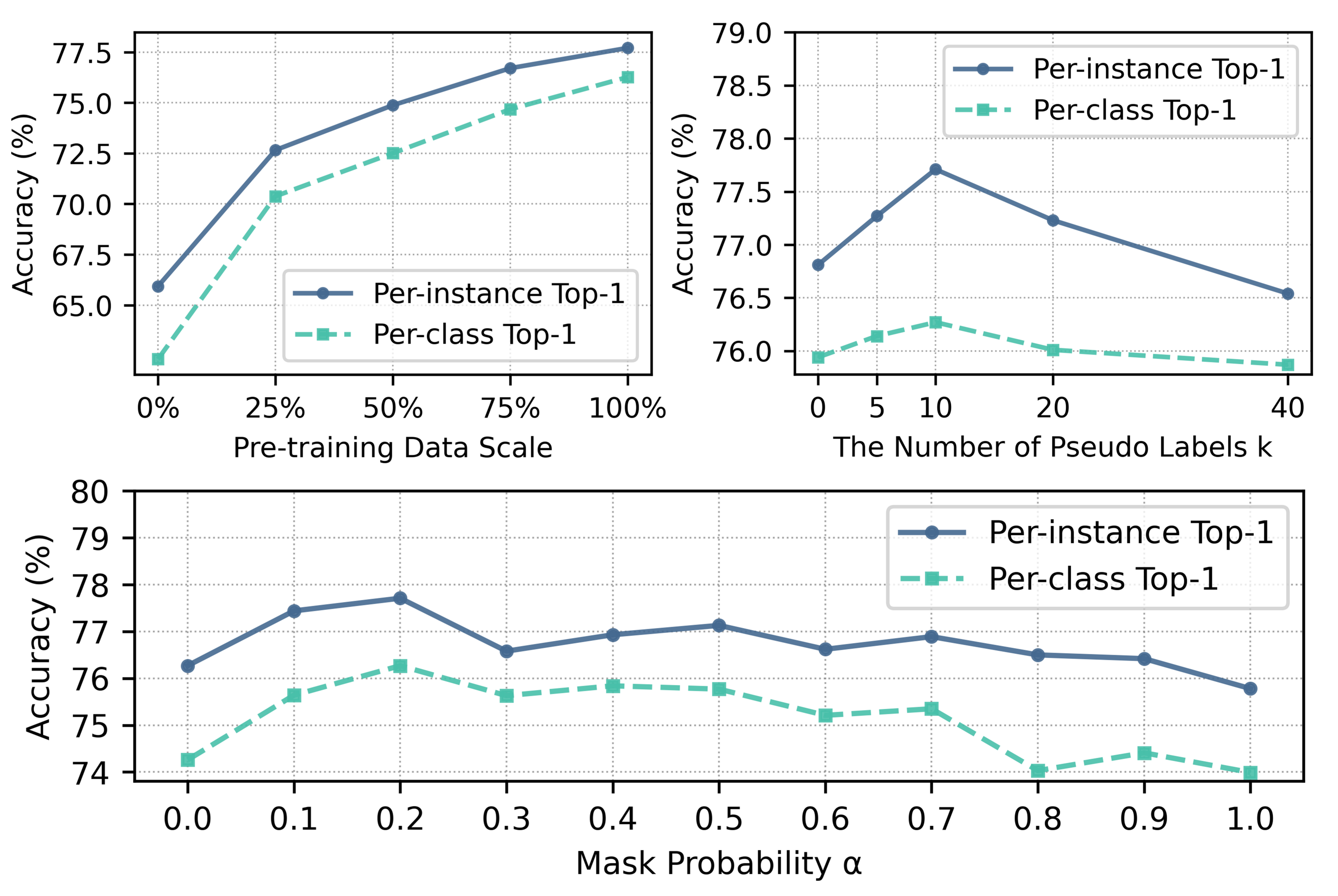}
    \caption{
    Hyperparameter selection on \modelname performance. 
    % Ablations of various hyper-parameters on CCL-SLR performance measured by accuracy (\%). 
    % We sequentially investigate the impact of changes in the pre-training data scale, the number of pseudo-positive labels $k$, and the mask probability $\alpha$ on CCL-SLR performance.
    We systematically investigate the impact of pre-training data scale, mask probability $\alpha$ in MPM, and number of pseudo-positive labels $k$ in SPM.
    }
    %\vspace{-1.5em}
    \label{fig:ab-figs}
\end{figure}

\noindent \textbf{Pre-Training Data Scale.}
% In Fig.~\ref{fig:ab-figs}, we explore the impact of the pre-training data scale on the performance of our proposed framework. 
% %The first row represents the scenario in which the framework is trained without pre-training. 
% The experimental results reveal that the performance exhibits a steady improvement as the proportion of the \text{pre-training} data scale increases, thereby indicating the scalability of our approach to large-scale sign language data.
In Fig.~\ref{fig:ab-figs}, we examine the impact of the pre-training data scale on the performance of our proposed framework.
The experimental results reveal that the performance exhibits a steady improvement as the proportion of the pre-training data scale increases, demonstrating that our model can effectively leverage large-scale unlabeled multi-modal sign language data to enhance its representation capability, highlighting the scalability of our approach to large-scale sign language data.

\noindent \textbf{Motion-Preserving Masking.} 
The influence of MPM is explored by sequentially changing the mask probability parameter $\alpha$ within $[0.0, 1.0]$. As shown in Fig.\ref{fig:ab-figs}, the performance reaches its peak at $\alpha = 0.2$, indicating that moderate masking effectively mitigates static bias while preserving essential dynamic information. However, an excessively high mask ratio can lead to performance degradation. Visualization results of MPM are provided in the Appendix.
%Appendix~\uppercase\expandafter{\romannumeral1}-B.

\noindent \textbf{Semantic Positive Mining.} 
% We examine how different $k$ values (the number of pseudo-positive labels in each branch) affect the performance. 
% We investigate the sensitivity of hyperparameter $k$, which denotes the number of mined semantic positives per modality.
% % As shown in Fig.~\ref{fig:ab-figs}, a too-small and a too-large $k$ value both lead to a performance drop. 
% As illustrated in Fig.~\ref{fig:ab-figs}, experimental results demonstrate that either extremely small or large values of $k$ lead to performance degradation.
% Based on this observation, we choose $k=10$ as default. 
We investigate the sensitivity of hyperparameter $k$ to our method, which denotes the number of mined semantic positives per modality. As illustrated in Fig.~\ref{fig:ab-figs}, experimental results demonstrate that either extremely small or large values of $k$ lead to performance degradation. An overly large $k$ will introduce noise during contrastive learning, negatively impacting the representation capability of the model. Based on these empirical findings, we set $k=10$ as the default value.
\begin{table}[!t]
\centering
\begin{tabular}{l | cc | cc}
\toprule[1.0pt]
  \multirow{2}{*}{Setting} & \multicolumn{2}{c|}{Per-instance} & \multicolumn{2}{c}{Per-class}\\
  & Top-1 & Top-5 & Top-1 & Top-5  \\
\midrule
 $\mathcal{M}_{f}$ only & 76.94 & 92.01 & 75.86 & 92.02  \\
 $\mathcal{M}_{g}$ and $\mathcal{M}_{f}$  & \textbf{77.71} & \textbf{92.28} & \textbf{76.27} & \textbf{92.30} \\
\bottomrule[1.0pt] 
\end{tabular}
\caption{Ablation studies on cross-modal contrastive learning loss.}
\label{tab:ab-memory-bank}
%\vspace{-1.5em}
\end{table} 
% \begin{table}[!t]
% \caption{Ablation studies for different pre-training modalities. 
% % ``Only Pose'' and ``Only RGB'' denotes that our framework is pre-trained only with Pose and RGB modality, respectively.
% }
% \label{tab:ab-diff-mod}
% \centering
% \begin{tabular}{cc | cc | cc}
% \toprule[1.0pt]
% \multirow{2}{*}{Pose modality} &\multirow{2}{*}{RGB modality} & \multicolumn{2}{c|}{Per-instance} & \multicolumn{2}{c}{Per-class} \\
%  & & Top-1 & Top-5 & Top-1 & Top-5 \\
% \midrule
% \ding{52} &  & 68.98 & 86.13 & 66.04 & 85.60 \\
%  & \ding{52} & 64.14 & 82.78 & 64.14 & 82.38 \\
% \ding{52} & \ding{52} & \textbf{77.71} & \textbf{92.28} & \textbf{76.27} & \textbf{92.30} \\
% \bottomrule[1.0pt]
% \end{tabular}
% \end{table}

\begin{table}[!t]
\centering
\resizebox{\linewidth}{!}{
\begin{tabular}{l | cc | cc | cc}
\toprule[1.0pt]
\multirow{2}{*}{Setting} & \multicolumn{2}{c|}{Fine-tuning} & \multicolumn{2}{c|}{Per-instance} & \multicolumn{2}{c}{Per-class} \\
& Pose & RGB & Top-1 & Top-5 & Top-1 & Top-5 \\
\midrule
\multirow{3}{*}{Baseline} & \Pmodal & 62.96 & 83.13 & 60.48 & 82.60 \\
& \Rmodal  & 60.25 & 82.78 & 57.16 & 80.35 \\
& \RPmodal & 70.04 & 88.43 & 67.73 & 86.91 \\
\midrule
\multirow{3}{*}{CCL-SLR} & \Pmodal & 71.98 & 89.93 & 70.04 & 89.41 \\
& \Rmodal  & 68.14 & 86.58 & 65.43 & 86.21 \\
& \RPmodal & \textbf{77.71} & \textbf{92.28} & \textbf{76.27} & \textbf{92.30} \\
\bottomrule[1.0pt]
\end{tabular}
}
\caption{Ablation studies on different pre-training modalities. 
% ``Only Pose'' and ``Only RGB'' denotes that our framework is pre-trained only with Pose and RGB modality, respectively.
}
\label{tab:ab-diff-mod}
%\vspace{-15pt}
\end{table}
\begin{figure}[t]
    \centering
    \includegraphics[width=\linewidth]{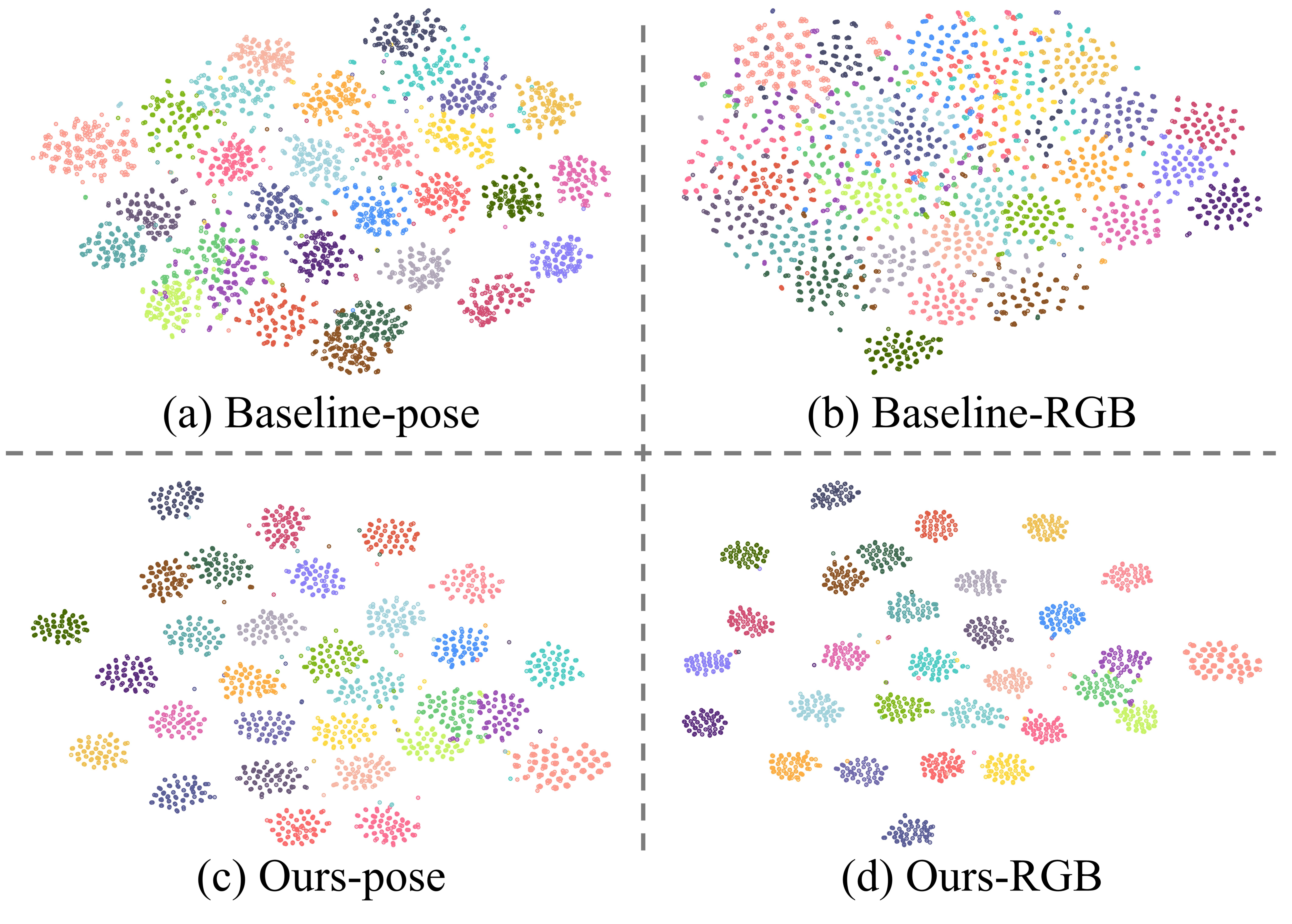}
    \caption{t-SNE visualizations of feature embeddings. We sample 30 sign words from the SLR500 dataset and visualize the feature embeddings extracted by CCL-SLR and its baseline, respectively.}
    \vspace{-1.5em}
    \label{fig:tsne}
\end{figure}

\noindent \textbf{Memory Banks.}
Unlike the vanilla MoCo v2~\cite{mocov2} framework, our pipeline maintains two memory banks ($\mathcal{M}_g$ and $\mathcal{M}_f$) in our pipeline: one for pseudo-labeling and the other for contrastive learning. Intuitively, maintaining a unique memory bank is structurally more concise. However, according to previous work~\cite{gupta2022understanding}, features before projection are more generally applicable, while features after projection are specialized for the instance discrimination task in contrastive learning. As shown in the Table~\ref{tab:ab-memory-bank}, our experimental results further confirm that directly using $\mathcal{M}_f$ for pseudo-labeling leads to a slight performance degradation.

\noindent \textbf{Cross-Modal Consistency Pre-training.} As illustrated in Table~\ref{tab:ab-diff-mod}, our framework is pre-trained under two different settings: baseline and CCL-SLR. The baseline conducts pre-training on two single-modal branches without key components of CCL-SLR. Compared with baseline, our CCL-SLR significantly improves the ISLR performance on the MSASL dataset for both single-modal and fusion results, which validates that our method effectively learns consistency representation across modalities. To further validate the effectiveness of our cross-modal consistency learning, t-SNE visualizations of the feature embeddings learned from \text{pre-training} are shown in Fig.~\ref{fig:tsne}.  
For both modalities, the representations learned by our CCL-SLR demonstrate more compact clustering patterns compared to the baseline method, indicating the superior representation learning capability of CCL-SLR.
\section{Conclusion}
% In this paper, we investigate cross-modal consistency learning for sign language recognition by leveraging complementary information from multi-modal data. We propose CCL-SLR, a novel multi-modal self-supervised pre-training framework that enhances sign representation through single- and cross-modal contrastive learning.
% To fully exploit the potential of multi-modal data, we introduce two techniques: MPM which enables visual encoders to capture motion-aware features from RGB videos, and SPM which addresses the issue of ``semantic collision'' in single-modal contrastive learning. 
% Extensive experiments demonstrate that our CCL-SLR achieves state-of-the-art performance across multiple benchmarks.
In this paper, we investigate the challenge of sign language recognition through cross-modal consistency learning. We propose CCL-SLR, a novel self-supervised multi-modal pre-training framework that enhances sign representation by leveraging consistency information across RGB and pose modalities.
Additionally, our framework incorporates two key innovations: MPM, which enables motion-aware regions extraction from RGB videos to tackle the problem of redundant features, and SPM, which effectively addresses the semantic collision problem in contrastive learning. 
% Extensive experiments demonstrate that our CCL-SLR achieves state-of-the-art performance across multiple benchmarks.
We conduct comprehensive evaluations on four widely used ISLR datasets, where CCL-SLR consistently achieves impressive performance, thereby validating the effectiveness of our proposed approach.

\clearpage
{
    \small
    \bibliographystyle{ieeenat_fullname}
    \bibliography{main}

\begin{thebibliography}{55}
\providecommand{\natexlab}[1]{#1}
\providecommand{\url}[1]{\texttt{#1}}
\expandafter\ifx\csname urlstyle\endcsname\relax
  \providecommand{\doi}[1]{doi: #1}\else
  \providecommand{\doi}{doi: \begingroup \urlstyle{rm}\Url}\fi

\bibitem[Carreira and Zisserman(2017)]{i3d}
Joao Carreira and Andrew Zisserman.
\newblock Quo vadis, action recognition? a new model and the kinetics dataset.
\newblock In \emph{CVPR}, pages 6299--6308, 2017.

\bibitem[Chen et~al.(2020{\natexlab{a}})Chen, Kornblith, Swersky, Norouzi, and Hinton]{simclrv2}
Ting Chen, Simon Kornblith, Kevin Swersky, Mohammad Norouzi, and Geoffrey~E Hinton.
\newblock Big self-supervised models are strong semi-supervised learners.
\newblock \emph{NeurIPS}, pages 22243--22255, 2020{\natexlab{a}}.

\bibitem[Chen et~al.(2020{\natexlab{b}})Chen, Fan, Girshick, and He]{mocov2}
Xinlei Chen, Haoqi Fan, Ross Girshick, and Kaiming He.
\newblock Improved baselines with momentum contrastive learning.
\newblock \emph{arxiv}, 2020{\natexlab{b}}.

\bibitem[Contributors(2020)]{mmpose2020}
MMPose Contributors.
\newblock {OpenMMLab Pose Estimation Toolbox and Benchmark}.
\newblock \url{https://github.com/open-mmlab/mmpose}, 2020.

\bibitem[Ding et~al.(2022)Ding, Li, Yang, Qian, Xu, Chen, Wang, and Xiong]{ding2022motion}
Shuangrui Ding, Maomao Li, Tianyu Yang, Rui Qian, Haohang Xu, Qingyi Chen, Jue Wang, and Hongkai Xiong.
\newblock Motion-aware contrastive video representation learning via foreground-background merging.
\newblock In \emph{CVPR}, pages 9716--9726, 2022.

\bibitem[Farhadi et~al.(2007)Farhadi, Forsyth, and White]{farhadi2007transfer}
Ali Farhadi, David Forsyth, and Ryan White.
\newblock Transfer learning in sign language.
\newblock In \emph{CVPR}, pages 1--8, 2007.

\bibitem[Fillbrandt et~al.(2003)Fillbrandt, Akyol, and Kraiss]{fillbrandt2003extraction}
Holger Fillbrandt, Suat Akyol, and K-F Kraiss.
\newblock Extraction of 3d hand shape and posture from image sequences for sign language recognition.
\newblock In \emph{2003 IEEE International SOI Conference. Proceedings (Cat. No. 03CH37443)}, pages 181--186, 2003.

\bibitem[Grill et~al.(2020)Grill, Strub, Altch{\'e}, Tallec, Richemond, Buchatskaya, Doersch, Avila~Pires, Guo, Gheshlaghi~Azar, et~al.]{grill2020bootstrap}
Jean-Bastien Grill, Florian Strub, Florent Altch{\'e}, Corentin Tallec, Pierre Richemond, Elena Buchatskaya, Carl Doersch, Bernardo Avila~Pires, Zhaohan Guo, Mohammad Gheshlaghi~Azar, et~al.
\newblock Bootstrap your own latent-a new approach to self-supervised learning.
\newblock \emph{NeurIPS}, pages 21271--21284, 2020.

\bibitem[Gupta et~al.(2022)Gupta, Ajanthan, Hengel, and Gould]{gupta2022understanding}
Kartik Gupta, Thalaiyasingam Ajanthan, Anton van~den Hengel, and Stephen Gould.
\newblock Understanding and improving the role of projection head in self-supervised learning.
\newblock \emph{arxiv}, 2022.

\bibitem[Hara et~al.(2017)Hara, Kataoka, and Satoh]{r3d}
Kensho Hara, Hirokatsu Kataoka, and Yutaka Satoh.
\newblock Learning spatio-temporal features with 3d residual networks for action recognition.
\newblock In \emph{ICCV}, pages 3154--3160, 2017.

\bibitem[He et~al.(2020)He, Fan, Wu, Xie, and Girshick]{he2020momentum}
Kaiming He, Haoqi Fan, Yuxin Wu, Saining Xie, and Ross Girshick.
\newblock Momentum contrast for unsupervised visual representation learning.
\newblock In \emph{CVPR}, pages 9729--9738, 2020.

\bibitem[Hinton(2015)]{hinton2015distilling}
Geoffrey Hinton.
\newblock Distilling the knowledge in a neural network.
\newblock \emph{arxiv}, 2015.

\bibitem[Hu et~al.(2021{\natexlab{a}})Hu, Zhao, Zhou, Wang, and Li]{signbert}
Hezhen Hu, Weichao Zhao, Wengang Zhou, Yuechen Wang, and Houqiang Li.
\newblock Signbert: Pre-training of hand-model-aware representation for sign language recognition.
\newblock In \emph{ICCV}, pages 11087--11096, 2021{\natexlab{a}}.

\bibitem[Hu et~al.(2021{\natexlab{b}})Hu, Zhou, and Li]{hma}
Hezhen Hu, Wengang Zhou, and Houqiang Li.
\newblock Hand-model-aware sign language recognition.
\newblock In \emph{AAAI}, pages 1558--1566, 2021{\natexlab{b}}.

\bibitem[Hu et~al.(2021{\natexlab{c}})Hu, Zhou, Pu, and Li]{nmfs}
Hezhen Hu, Wengang Zhou, Junfu Pu, and Houqiang Li.
\newblock Global-local enhancement network for nmf-aware sign language recognition.
\newblock \emph{ACM TOMM}, 17\penalty0 (3):\penalty0 1--19, 2021{\natexlab{c}}.

\bibitem[Hu et~al.(2023)Hu, Zhao, Zhou, and Li]{signbert+}
Hezhen Hu, Weichao Zhao, Wengang Zhou, and Houqiang Li.
\newblock Signbert+: Hand-model-aware self-supervised pre-training for sign language understanding.
\newblock \emph{IEEE TPAMI}, 45\penalty0 (9):\penalty0 11221--11239, 2023.

\bibitem[Huang et~al.(2015)Huang, Zhou, Li, and Li]{hj}
Jie Huang, Wengang Zhou, Houqiang Li, and Weiping Li.
\newblock Sign language recognition using 3d convolutional neural networks.
\newblock In \emph{ICME}, pages 1--6. IEEE, 2015.

\bibitem[Huang et~al.(2018)Huang, Zhou, Li, and Li]{slr500}
Jie Huang, Wengang Zhou, Houqiang Li, and Weiping Li.
\newblock {Attention-based 3D-CNNs for large-vocabulary sign language recognition}.
\newblock \emph{IEEE TCSVT}, 29\penalty0 (9):\penalty0 2822--2832, 2018.

\bibitem[Jeon et~al.(2021)Jeon, Min, Kim, and Sohn]{jeon2021mining}
Sangryul Jeon, Dongbo Min, Seungryong Kim, and Kwanghoon Sohn.
\newblock Mining better samples for contrastive learning of temporal correspondence.
\newblock In \emph{CVPR}, pages 1034--1044, 2021.

\bibitem[Jiang et~al.(2021{\natexlab{a}})Jiang, Sun, Wang, Bai, Li, and Fu]{jiang2021skeleton}
Songyao Jiang, Bin Sun, Lichen Wang, Yue Bai, Kunpeng Li, and Yun Fu.
\newblock Skeleton aware multi-modal sign language recognition.
\newblock In \emph{CVPR}, pages 3413--3423, 2021{\natexlab{a}}.

\bibitem[Jiang et~al.(2021{\natexlab{b}})Jiang, Sun, Wang, Bai, Li, and Fu]{sam}
Songyao Jiang, Bin Sun, Lichen Wang, Yue Bai, Kunpeng Li, and Yun Fu.
\newblock Sign language recognition via skeleton-aware multi-model ensemble.
\newblock \emph{arxiv}, 2021{\natexlab{b}}.

\bibitem[Jiang et~al.(2023)Jiang, Lu, Zhang, Ma, Han, Lyu, Li, and Chen]{Jiang2023RTMPoseRM}
Tao Jiang, Peng Lu, Li Zhang, Ning Ma, Rui Han, Chengqi Lyu, Yining Li, and Kai Chen.
\newblock {RTMPose}: Real-time multi-person pose estimation based on mmpose.
\newblock \emph{arxiv}, 2023.

\bibitem[Jin et~al.(2020)Jin, Xu, Xu, Wang, Liu, Qian, Ouyang, and Luo]{jin2020whole}
Sheng Jin, Lumin Xu, Jin Xu, Can Wang, Wentao Liu, Chen Qian, Wanli Ouyang, and Ping Luo.
\newblock Whole-body human pose estimation in the wild.
\newblock In \emph{ECCV}, 2020.

\bibitem[Joze and Koller(2018)]{msasl}
Hamid Reza~Vaezi Joze and Oscar Koller.
\newblock Ms-asl: A large-scale data set and benchmark for understanding american sign language.
\newblock \emph{arxiv}, 2018.

\bibitem[Koller et~al.(2018)Koller, Zargaran, Ney, and Bowden]{koller2018deep}
Oscar Koller, Sepehr Zargaran, Hermann Ney, and Richard Bowden.
\newblock Deep sign: Enabling robust statistical continuous sign language recognition via hybrid cnn-hmms.
\newblock \emph{IJCV}, 126:\penalty0 1311--1325, 2018.

\bibitem[Li et~al.(2020{\natexlab{a}})Li, Rodriguez, Yu, and Li]{wlasl}
Dongxu Li, Cristian Rodriguez, Xin Yu, and Hongdong Li.
\newblock Word-level deep sign language recognition from video: A new large-scale dataset and methods comparison.
\newblock In \emph{CVPR}, pages 1459--1469, 2020{\natexlab{a}}.

\bibitem[Li et~al.(2020{\natexlab{b}})Li, Yu, Xu, Petersson, and Li]{li2020transferring}
Dongxu Li, Xin Yu, Chenchen Xu, Lars Petersson, and Hongdong Li.
\newblock Transferring cross-domain knowledge for video sign language recognition.
\newblock In \emph{CVPR}, pages 6205--6214, 2020{\natexlab{b}}.

\bibitem[Li et~al.(2020{\natexlab{c}})Li, Yu, Xu, Petersson, and Li]{tck}
Dongxu Li, Xin Yu, Chenchen Xu, Lars Petersson, and Hongdong Li.
\newblock Transferring cross-domain knowledge for video sign language recognition.
\newblock In \emph{CVPR}, pages 6205--6214, 2020{\natexlab{c}}.

\bibitem[Li et~al.(2022)Li, Xu, Liu, Zhong, Wang, Petersson, and Li]{li2022transcribing}
Dongxu Li, Chenchen Xu, Liu Liu, Yiran Zhong, Rong Wang, Lars Petersson, and Hongdong Li.
\newblock Transcribing natural languages for the deaf via neural editing programs.
\newblock In \emph{AAAI}, pages 11991--11999, 2022.

\bibitem[Lin et~al.(2019)Lin, Gan, and Han]{lin2019tsm}
Ji Lin, Chuang Gan, and Song Han.
\newblock Tsm: Temporal shift module for efficient video understanding.
\newblock In \emph{ICCV}, pages 7083--7093, 2019.

\bibitem[Mohaghegh~Dolatabadi et~al.(2020)Mohaghegh~Dolatabadi, Erfani, and Leckie]{advflow}
Hadi Mohaghegh~Dolatabadi, Sarah Erfani, and Christopher Leckie.
\newblock Advflow: Inconspicuous black-box adversarial attacks using normalizing flows.
\newblock \emph{NeurIPS}, pages 15871--15884, 2020.

\bibitem[Ong and Ranganath(2005)]{ong2005automatic}
Sylvie~CW Ong and Surendra Ranganath.
\newblock Automatic sign language analysis: A survey and the future beyond lexical meaning.
\newblock \emph{IEEE Trans. Pattern Anal. Mach. Intell.}, 27\penalty0 (06):\penalty0 873--891, 2005.

\bibitem[Radford et~al.(2021)Radford, Kim, Hallacy, Ramesh, Goh, Agarwal, Sastry, Askell, Mishkin, Clark, et~al.]{radford2021learning}
Alec Radford, Jong~Wook Kim, Chris Hallacy, Aditya Ramesh, Gabriel Goh, Sandhini Agarwal, Girish Sastry, Amanda Askell, Pamela Mishkin, Jack Clark, et~al.
\newblock Learning transferable visual models from natural language supervision.
\newblock In \emph{ICML}, pages 8748--8763, 2021.

\bibitem[Rastgoo et~al.(2021)Rastgoo, Kiani, and Escalera]{rastgoo2021sign}
Razieh Rastgoo, Kourosh Kiani, and Sergio Escalera.
\newblock Sign language recognition: A deep survey.
\newblock \emph{Expert Systems with Applications}, 164:\penalty0 113794, 2021.

\bibitem[Selvaraju et~al.(2017)Selvaraju, Cogswell, Das, Vedantam, Parikh, and Batra]{8237336}
Ramprasaath~R. Selvaraju, Michael Cogswell, Abhishek Das, Ramakrishna Vedantam, Devi Parikh, and Dhruv Batra.
\newblock Grad-cam: Visual explanations from deep networks via gradient-based localization.
\newblock In \emph{ICCV}, pages 618--626, 2017.

\bibitem[Shen et~al.(2024)Shen, Zheng, and Yang]{shen2024stepnet}
Xiaolong Shen, Zhedong Zheng, and Yi Yang.
\newblock Stepnet: Spatial-temporal part-aware network for isolated sign language recognition.
\newblock \emph{ACM TOMM}, 20\penalty0 (7):\penalty0 1--19, 2024.

\bibitem[Starner(1995)]{starner1995visual}
Thad Starner.
\newblock \emph{Visual recognition of american sign language using hidden markov models}.
\newblock PhD thesis, Massachusetts Institute of Technology, 1995.

\bibitem[Wang et~al.(2022)Wang, Bertasius, Tran, and Torresani]{wang2022long}
Jue Wang, Gedas Bertasius, Du Tran, and Lorenzo Torresani.
\newblock Long-short temporal contrastive learning of video transformers.
\newblock In \emph{CVPR}, pages 14010--14020, 2022.

\bibitem[Xu et~al.(2018)Xu, Yang, Fan, Yue, Liang, Yang, and Huang]{infonce}
Ning Xu, Linjie Yang, Yuchen Fan, Dingcheng Yue, Yuchen Liang, Jianchao Yang, and Thomas Huang.
\newblock Youtube-vos: A large-scale video object segmentation benchmark.
\newblock \emph{arxiv}, 2018.

\bibitem[Yan et~al.(2018{\natexlab{a}})Yan, Xiong, and Lin]{stgcn}
Sijie Yan, Yuanjun Xiong, and Dahua Lin.
\newblock Spatial temporal graph convolutional networks for skeleton-based action recognition.
\newblock In \emph{AAAI}, 2018{\natexlab{a}}.

\bibitem[Yan et~al.(2018{\natexlab{b}})Yan, Xiong, and Lin]{yan2018stgcn}
Sijie Yan, Yuanjun Xiong, and Dahua Lin.
\newblock Spatial temporal graph convolutional networks for skeleton-based action recognition.
\newblock In \emph{AAAI}, pages 1--10, 2018{\natexlab{b}}.

\bibitem[Zhang et~al.(2022{\natexlab{a}})Zhang, Hou, Zhang, and Li]{zhang2022contrastive}
Haoyuan Zhang, Yonghong Hou, Wenjing Zhang, and Wanqing Li.
\newblock Contrastive positive mining for unsupervised 3d action representation learning.
\newblock In \emph{ECCV}, pages 36--51, 2022{\natexlab{a}}.

\bibitem[Zhang et~al.(2022{\natexlab{b}})Zhang, Li, Liu, Zhang, Su, Zhu, Ni, and Shum]{zhang2022dino}
Hao Zhang, Feng Li, Shilong Liu, Lei Zhang, Hang Su, Jun Zhu, Lionel~M Ni, and Heung-Yeung Shum.
\newblock Dino: Detr with improved denoising anchor boxes for end-to-end object detection.
\newblock \emph{arXiv}, 2022{\natexlab{b}}.

\bibitem[Zhang et~al.(2016)Zhang, Zhou, Xie, Pu, and Li]{zjh}
Jihai Zhang, Wengang Zhou, Chao Xie, Junfu Pu, and Houqiang Li.
\newblock Chinese sign language recognition with adaptive hmm.
\newblock In \emph{ICME}, pages 1--6. IEEE, 2016.

\bibitem[Zhang et~al.(2022{\natexlab{c}})Zhang, Wang, and Ma]{ssvc}
Manlin Zhang, Jinpeng Wang, and Andy~J Ma.
\newblock Suppressing static visual cues via normalizing flows for self-supervised video representation learning.
\newblock In \emph{AAAI}, pages 3300--3308, 2022{\natexlab{c}}.

\bibitem[Zhao et~al.(2023)Zhao, Hu, Zhou, Shi, and Li]{best}
Weichao Zhao, Hezhen Hu, Wengang Zhou, Jiaxin Shi, and Houqiang Li.
\newblock Best: Bert pre-training for sign language recognition with coupling tokenization.
\newblock In \emph{AAAI}, pages 3597--3605, 2023.

\bibitem[Zhao et~al.(2024{\natexlab{a}})Zhao, Hu, Zhou, Mao, Wang, and Li]{masa}
Weichao Zhao, Hezhen Hu, Wengang Zhou, Yunyao Mao, Min Wang, and Houqiang Li.
\newblock Masa: Motion-aware masked autoencoder with semantic alignment for sign language recognition.
\newblock \emph{IEEE TCSVT}, 34\penalty0 (11):\penalty0 10793--10804, 2024{\natexlab{a}}.

\bibitem[Zhao et~al.(2024{\natexlab{b}})Zhao, Zhou, Hu, Wang, and Li]{stc-slr}
Weichao Zhao, Wengang Zhou, Hezhen Hu, Min Wang, and Houqiang Li.
\newblock Self-supervised representation learning with spatial-temporal consistency for sign language recognition.
\newblock \emph{arxiv}, 2024{\natexlab{b}}.

\bibitem[Zhao et~al.(2024{\natexlab{c}})Zhao, Zhou, Hu, Wang, and Li]{zhao2024self}
Weichao Zhao, Wengang Zhou, Hezhen Hu, Min Wang, and Houqiang Li.
\newblock Self-supervised representation learning with spatial-temporal consistency for sign language recognition.
\newblock \emph{IEEE TIP}, 33:\penalty0 4188--4201, 2024{\natexlab{c}}.

\bibitem[Zhou et~al.(2024)Zhou, Zhao, Hu, Li, and Li]{zhou2024scaling}
Wengang Zhou, Weichao Zhao, Hezhen Hu, Zecheng Li, and Houqiang Li.
\newblock Scaling up multimodal pre-training for sign language understanding.
\newblock \emph{arXiv preprint arXiv:2408.08544}, 2024.

\bibitem[Zhu et~al.(2017)Zhu, Li, Ouyang, Yu, and Wang]{zhu2017learning}
Feng Zhu, Hongsheng Li, Wanli Ouyang, Nenghai Yu, and Xiaogang Wang.
\newblock Learning spatial regularization with image-level supervisions for multi-label image classification.
\newblock In \emph{CVPR}, pages 5513--5522, 2017.

\bibitem[Zhu et~al.(2023{\natexlab{a}})Zhu, Fu, and Wu]{multilabel}
Ke Zhu, Minghao Fu, and Jianxin Wu.
\newblock Multi-label self-supervised learning with scene images.
\newblock In \emph{ICCV}, pages 6694--6703, 2023{\natexlab{a}}.

\bibitem[Zhu et~al.(2023{\natexlab{b}})Zhu, Han, Yu, and Liu]{zhu2023modeling}
Yisheng Zhu, Hu Han, Zhengtao Yu, and Guangcan Liu.
\newblock Modeling the relative visual tempo for self-supervised skeleton-based action recognition.
\newblock In \emph{ICCV}, pages 13913--13922, 2023{\natexlab{b}}.

\bibitem[Zolfaghari et~al.(2021)Zolfaghari, Zhu, Gehler, and Brox]{crossclr}
Mohammadreza Zolfaghari, Yi Zhu, Peter Gehler, and Thomas Brox.
\newblock Crossclr: Cross-modal contrastive learning for multi-modal video representations.
\newblock In \emph{ICCV}, pages 1450--1459, 2021.

\bibitem[Zuo et~al.(2023)Zuo, Wei, and Mak]{nla-slr}
Ronglai Zuo, Fangyun Wei, and Brian Mak.
\newblock Natural language-assisted sign language recognition.
\newblock In \emph{CVPR}, pages 14890--14900, 2023.

\end{thebibliography}
}

\clearpage
\maketitlesupplementary
\setcounter{section}{0}
\renewcommand\thesection{\Alph{section}}
\renewcommand\thefigure{\arabic{figure}}
\renewcommand\thetable{\arabic{table}}
\renewcommand\theequation{\arabic{equation}}
% \renewcommand\thesection{\Alph{section}}

% \section{Rationale}
% \label{sec:rationale}
% % 
% Having the supplementary compiled together with the main paper means that:
% % 
% \begin{itemize}
% \item The supplementary can back-reference sections of the main paper, for example, we can refer to \cref{sec:intro};
% \item The main paper can forward reference sub-sections within the supplementary explicitly (e.g. referring to a particular experiment); 
% \item When submitted to arXiv, the supplementary will already included at the end of the paper.
% \end{itemize}
% % 
% To split the supplementary pages from the main paper, you can use \href{https://support.apple.com/en-ca/guide/preview/prvw11793/mac#:~:text=Delete%20a%20page%20from%20a,or%20choose%20Edit%20%3E%20Delete).}{Preview (on macOS)}, \href{https://www.adobe.com/acrobat/how-to/delete-pages-from-pdf.html#:~:text=Choose%20%E2%80%9CTools%E2%80%9D%20%3E%20%E2%80%9COrganize,or%20pages%20from%20the%20file.}{Adobe Acrobat} (on all OSs), as well as \href{https://superuser.com/questions/517986/is-it-possible-to-delete-some-pages-of-a-pdf-document}{command line tools}.

\section{Framework Implementation}
\label{sec:sup_1}
In this section, we describe the details of the pose data extraction, the encoder architecture in the pose branch and data preprocessing.

\subsection{Pose Keypoint Extraction.} 
\begin{figure}[t]
    \centering
    \includegraphics[width=\linewidth]{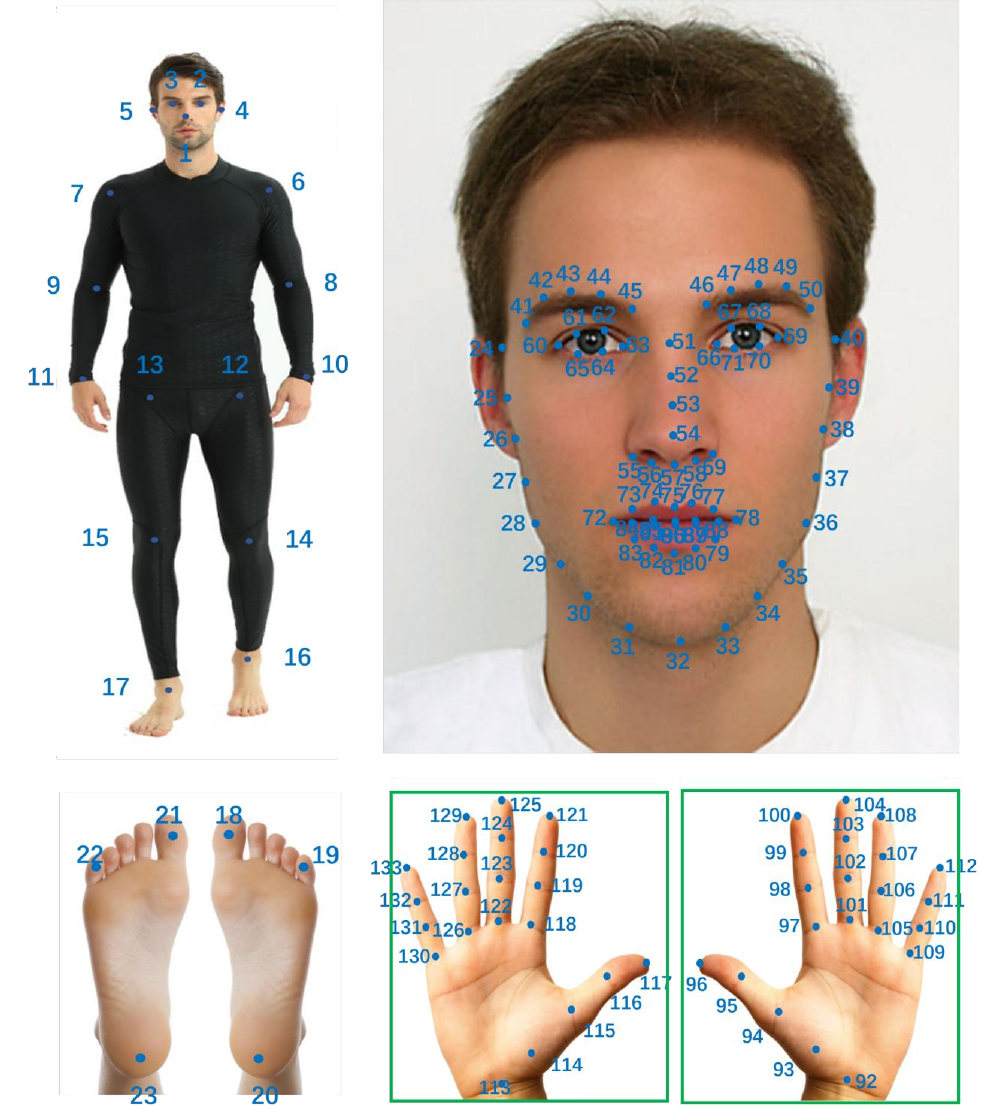}
    \caption{The visualization of the whole-body 133 keypoints from ~\cite{jin2020whole}.}
    \label{ap:coco_whole_body}
\end{figure}

We employ RTMPose-x~\cite{Jiang2023RTMPoseRM} from MMPose to extract 133 whole-body keypoints, which provide detailed spatial information for subsequent pose analysis.
The visualization of whole-body keypoints are shown in Figure~\ref{ap:coco_whole_body}.
We select 75 keypoints and divide the keypoints into several sub-pose (left hand, right hand, face, mouth, and body).
We select the indices for the left hand (\{92-112\}), right hand (\{113-133\}), face (\{24-41, 54\}), mouth (\{84-92\}) and body (\{0,5,7,9,6,8,10\}) to represent each group (denoted as $\mathcal{G}_r, r\in\{LH,RH,F,M,B\}$).

\subsection{Encoder of Pose Branch.} We utilize a GCN+Transformer architecture~\cite{zhou2024scaling} as the pose branch encoder. Specifically, four ST-GCN modules~\cite{yan2018stgcn} are employed to extract group-specific features $\mathbf{v}_r$ from pose group $\mathcal{G}_r$, where the left and right hands share the same module for feature extraction. The extracted 512 dimension features are then concatenated into manual features $\mathbf{v}_{man}$ and non-manual features $\mathbf{v}_{non}$:
\begin{align}
\mathbf{v}_{man} &= \text{Concat}([\mathbf{v}_{LH}, \mathbf{v}_{RH}, \mathbf{v}_{B}]),\\
\mathbf{v}_{non} &= \text{Concat}([\mathbf{v}_{F}, \mathbf{v}_{M}]),
\end{align}
where $\mathbf{v}_{LH}$, $\mathbf{v}_{RH}$, $\mathbf{v}_{B}$, $\mathbf{v}_{F}$, and $\mathbf{v}_{M}$ represent features from left hand, right hand, body, face, and mouth, respectively. 
To model temporal relationships, two separate 3-block 8-head Transformers ($\mathcal{T}_{man}$ and $\mathcal{T}_{non}$) are leveraged to process these features and generate the final pose embedding:
\begin{equation}
\mathbf{v} = \text{Concat}([\mathcal{T}_{man}(\mathbf{v}_{man}), \mathcal{T}_{non}(\mathbf{v}_{non})]),
\end{equation}
where final pose embedding $\mathbf{v}$ is 1024 dimensions.
% Hyperparameters of our used transformer are shown in Tab.~\ref{tab:Feature Extraction}.
\subsection{Data Preprocessing.} 
For data preprocessing, we select aforementioned 75 keypoints for pose input per frame and resize all RGB frames to $224 \times 224$. 
For computational efficiency, we uniformly sample $T = 32$ frames from each sequence. 

\subsection{Data Augmentation}
For the query and key samples, we ensure consistency in hyperparameters during data augmentation while performing the process independently. For random temporal cropping, we first apply continuous temporal cropping to the input sequence, randomly extracting a clip from the interval $[lT, T]$ frames. Subsequently, we uniformly sample a fixed number of  $K$ frames from the cropped interval. In our approach, we set $l = 0.1$ and $K$ = 64.

\section{Visualization of Motion-Preserving Masking}
\label{sec:sup_2}
\begin{figure*}[t]
    \centering
    \includegraphics[width=\linewidth]{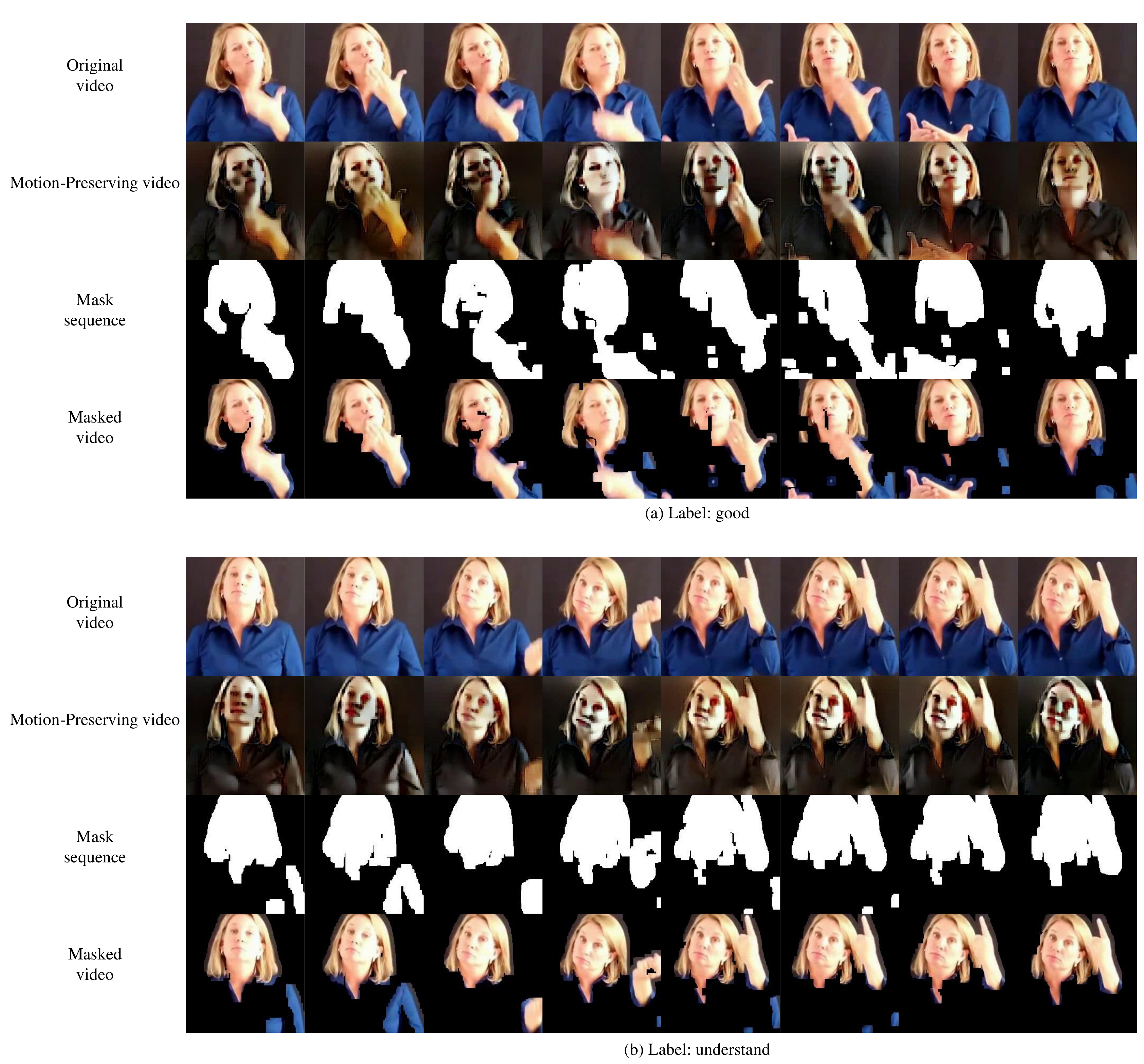}
    \caption{The visualizations of motion-preserving masking.}
    \label{sp:mpm1}
\end{figure*}

\begin{figure*}[t]
    \centering
    \includegraphics[width=\linewidth]{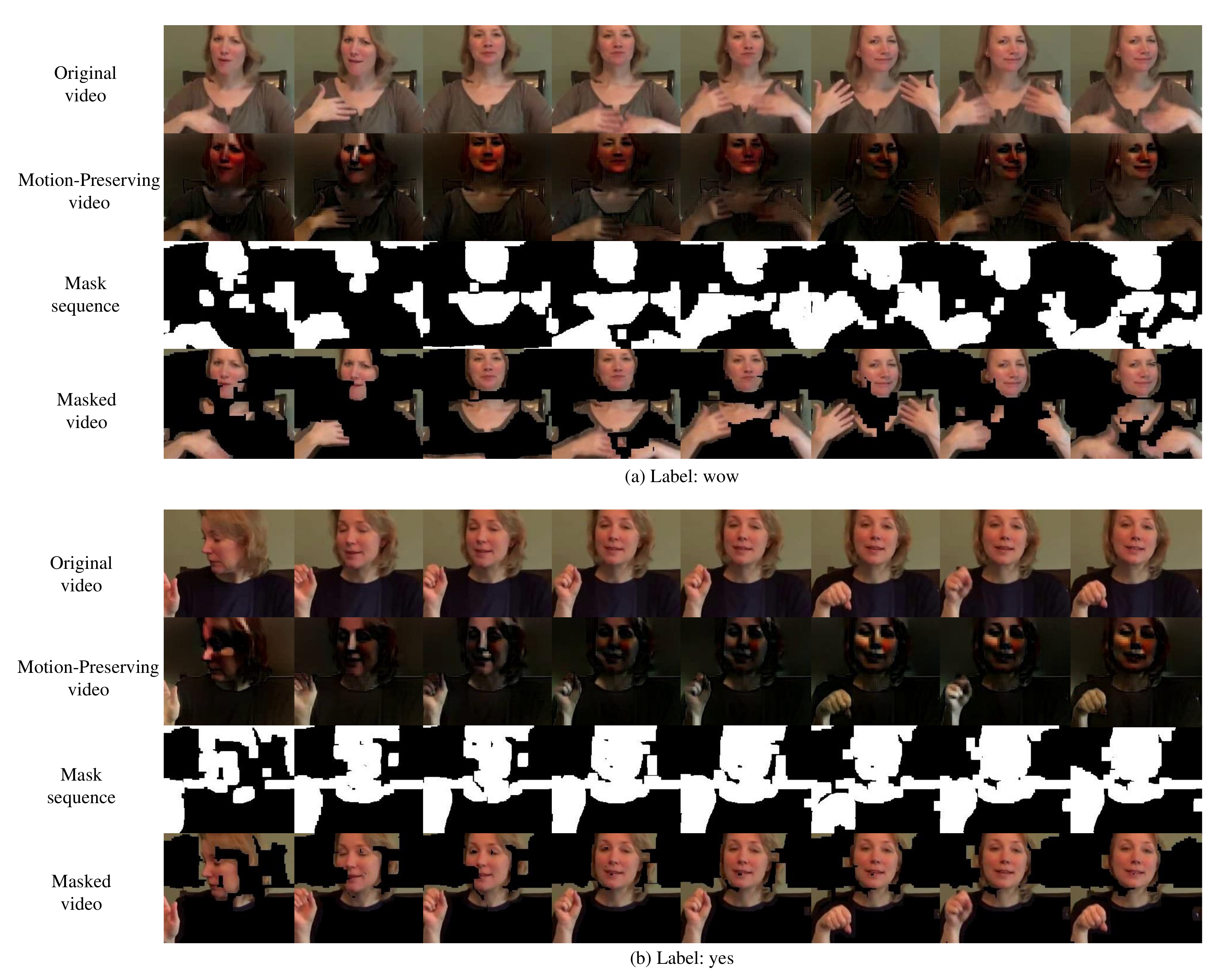}
    \caption{The visualizations of motion-preserving masking.}
    \label{sp:mpm2}
\end{figure*}

% In this section, we present the visualization results of Motion-Preserving Masking (MPM). 
As shown in Fig.~\ref{sp:mpm1} and Fig.~\ref{sp:mpm2}, the generated motion-preserving videos effectively suppress static regions while highlighting motion areas. 
Furthermore, since motion-preserving videos significantly alter the pixel distribution of the original videos, we further generate binary motion-preserving mask sequences from the motion-preserving videos. These mask sequence are applied to mask original videos, explicitly mitigating static information redundancy in RGB modality. The visualizations indicate that MPM preserves semantically informative regions such as hand and facial features, while effectively suppressing semantically irrelevant areas including clothing textures and background elements, enhancing the cross-modal feature alignment between pose and RGB representations.


\end{document}